\documentclass{article}
\usepackage[preprint]{neurips_2026}

\usepackage{microtype}
\usepackage{hyperref}
\usepackage{url}
\usepackage{booktabs}
\usepackage{graphicx}
\usepackage{caption}
\usepackage{subcaption}

\usepackage[dvipsnames]{xcolor}
\usepackage[breakable]{tcolorbox}

\usepackage{makecell}

\usepackage{threeparttable}

\usepackage{amsmath,amsfonts}
\usepackage{float}
\usepackage[ruled,vlined,linesnumbered]{algorithm2e}
\usepackage{amssymb}

\usepackage{ragged2e} 

\usepackage{colortbl}
\usepackage{multirow}

\definecolor{speedupgreen}{HTML}{1B7A3D}
\definecolor{speedupbg}{HTML}{E8F5E9}
\definecolor{accdrop}{HTML}{C62828}
\definecolor{accgain}{HTML}{1B5E20}
\definecolor{headerbg}{HTML}{1A237E}
\definecolor{headertext}{HTML}{FFFFFF}
\definecolor{lightgray}{HTML}{F5F5F5}
\definecolor{midgray}{HTML}{E0E0E0}

\newcommand{\speedup}[1]{\textcolor{speedupgreen}
{\textbf{#1$\times$}}}
\newcommand{\accdelta}[1]{
  \ifdim #1pt < 0pt
    \textcolor{accdrop}{\small #1}
  \else
    \textcolor{accgain}{\small +#1}
  \fi
}

\newcommand{\speeddown}[1]{\textcolor{accdrop}{\textbf{#1$\times$}}}

\usepackage[nameinlink,noabbrev]{cleveref}

\usepackage{lineno}

\definecolor{darkblue}{rgb}{0, 0, 0.5}
\hypersetup{colorlinks=true, citecolor=darkblue, linkcolor=darkblue, urlcolor=darkblue}

\usepackage{xspace}
\newcommand{\sys}{\textsc{Parse}\xspace}

\newcommand{\gcomment}[1]{\tcp{\textcolor{green!60!black}{$\triangleright$~#1}}}

\newcommand{\ginline}[1]{\tcp*[r]{\textcolor{green!60!black}{$\triangleright$~#1}}}

\title{Parallel Prefix Verification for Speculative Generation}

\author{%
  Yuncheng Yao \\
  Duke University \\
  \And
  Yuxuan Xia \\
  New York University \\
  \And
  Shengjie Wang \\
  New York University \\
  \And
  Danyang Zhuo \\
  Duke University \\
}

\begin{document}
\raggedbottom

\maketitle

\begin{abstract}

We introduce \sys (\underline{PA}rallel p\underline{R}efix \underline{S}peculative \underline{E}ngine), a speculative generation framework that accelerates large language model (LLM) inference by parallelizing prefix verification on a semantic level.
Existing speculative decoding methods are fundamentally limited by token-level equivalence: the target model must verify each token, leading to short acceptance lengths and modest speedups. Moving to semantic or segment-level verification can substantially increase acceptance granularity, but prior approaches rely on sequential verification, introducing significant overhead and limiting practical gains. 
\sys introduces parallel prefix verification, enabling semantic-level verification without sequential checks. Given a full draft from a draft model, the target model evaluates correctness across multiple prefixes in a single forward pass using a custom attention mask, directly identifying the maximal valid prefix. This eliminates sequential segment verification, and makes verification compute-efficient.
\sys is orthogonal to token-level speculative decoding and can be composed with it for additional gains. Across models and benchmarks, \sys delivers 1.25$\times$--4.3$\times$ throughput gain over the target model, and 1.6$\times$--4.5$\times$ when composed with EAGLE-3 --- all with negligible accuracy degradation. This demonstrates parallel prefix verification as an effective, general approach to accelerating LLM inference.

\end{abstract}

\section{Introduction}
As large language models (LLMs) scale, their inference cost increasingly dominates deployment budgets. Reducing latency and server load without compromising quality has therefore become a central research goal~\citep{pagedattention, orca, prefilldecode_diss}. Among existing techniques, \emph{speculative generation}~\citep{specdecode, specsample} is especially attractive: a lightweight draft model proposes tokens that the target model then verifies. Classical speculative \emph{decoding} is \emph{lossless}: through rejection sampling, the final output distribution exactly matches that of the target model, so any accepted tokens are guaranteed to be distributionally consistent.

While token-level speculation provides rigorous correctness, it constrains acceleration to per-token agreement. Many user-facing tasks admit a coarser but still reliable unit of verification: \emph{semantic spans}. In practice, correctness is often determined by whether a prefix remains on a correct solution trajectory (e.g., problem setup, derivation scaffolding, or factual context), not by matching the target model’s probability at each token position. This observation opens a complementary axis of speedup: amortize verification over longer semantic units, reuse any prefix that is semantically sound, and only regenerate from the first incorrect point onward. 

Prior works have also recognized the potential of semantic-level speculation. In particular, SpecReason~\citep{specreason} and Speculative Thinking~\citep{specthinking} apply semantic verification to reasoning-oriented tasks by checking intermediate chain-of-thought segments (e.g., a step in reasoning). However, these approaches operate at the segment level and rely on sequential verification along the generation trajectory. Each segment must be checked before proceeding to the next, introducing significant overhead and unable to fully unlock the benefits of larger acceptance units.

These observations expose a fundamental tension in speculative generation: \textbf{fine-grained (token-level) verification enables parallelism but limits acceptance length, whereas coarse-grained (semantic-level) verification increases acceptance length but introduces sequential dependency.}

In this paper, we show that this tradeoff is not inherent. We introduce parallel prefix verification, a new inference paradigm that enables semantic-level parallel prefix verification. Our key idea is to treat every partial output as a prefix of a complete draft and to verify multiple prefixes simultaneously in a single forward pass. Given a complete candidate output from a draft model, the target model evaluates correctness across different prefix boundaries using a custom attention mask, and identifies the maximal valid prefix. This eliminates iterative, segment-by-segment generation and verification.

Building on this idea, we present \sys, a speculative generation framework that combines a fast draft model with parallel prefix verification in the target model. \sys requires no model retraining and \emph{integrates cleanly with existing token-level speculative decoding techniques}. Empirically, using Qwen3-235B~\citep{qwen3} as the target model, \sys achieves $\sim$1.3$\times$–4.3$\times$ speedup while maintaining accuracy close to the target model, demonstrating that semantic-level parallel verification is a practical and effective approach for high-throughput LLM serving.

\begin{figure}[t]
    \centering
    \includegraphics[width=\linewidth]{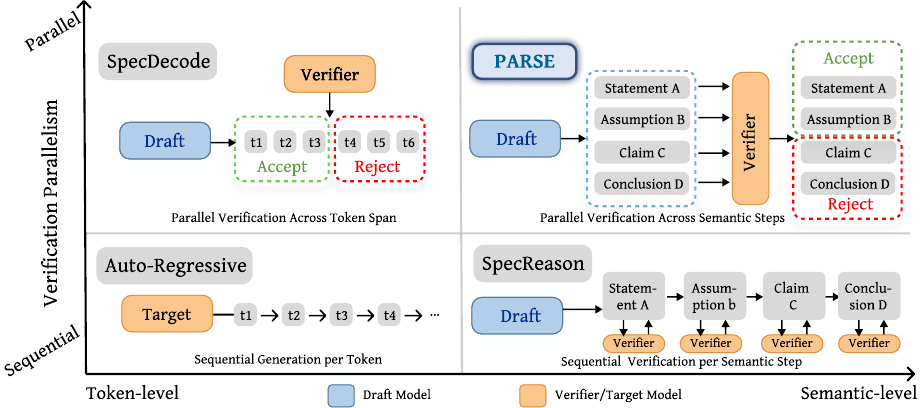}

    \caption{Auto-regressive decoding is sequential and token-level; SpecDecode parallelizes verification at the token level (short acceptance spans); SpecReason verifies semantic steps but sequentially (serial bottleneck). \sys fills the remaining quadrant by verifying all semantic prefixes of a draft in a single prefill and resuming from the last correct one, capturing both long acceptance units and parallel verification.}
    \label{fig:comparison}
\end{figure}

\section{Related Works}

\subsection{Speculative Decoding}

Speculative decoding has recently seen several enhancements to draft tree structures, feature‐level modeling, and single‐model drafting. The EAGLE series (EAGLE-1, EAGLE-2)~\citep{eagle1,eagle2} introduces draft trees, enabling context-aware token drafting with lossless speedups of 3--4.3$\times$. Dynamic Depth Decoding~\citep{brown2024dynamic} further improves on EAGLE-2 by adapting draft depth per context, gaining additional speed without accuracy loss. EAGLE-3~\citep{eagle3} relaxes the feature-prediction constraint by fusing multiple semantic feature levels during training and inference, achieving better speedups. Alternatives like Medusa~\citep{medusa2023} avoid separate small models by attaching lightweight “draft heads” directly to the target model.

These variants of speculative decoding still rely on \emph{token-level} distributional alignment, where the target model sequentially verifies agreement on each token. In contrast, \sys shifts the focus to \emph{semantic-level} verification: instead of matching token probabilities, the target model evaluates whether draft prefixes remain semantically correct. This is done in a single prefill pass, after which the target model continues from the last verified prefix. Figure~\ref{fig:comparison} illustrates this contrast between token-level speculative decoding and our approach.

\subsection{Semantic Speculative Decoding}

Prior works such as JudgeDecoding~\citep{judge_decoding} enables a larger acceptance length (about 19 tokens on average) by training a small network to recognize semantic-level alignment based on the last layer of network features. SpecReason~\citep{specreason}, Lookahead Reasoning~\citep{speclookahead} and Speculative Thinking~\citep{specthinking} apply semantic verification to reasoning steps in a chain-of-thought (CoT), enabling speed-up in the reasoning process. In these approaches, a small draft model proposes an intermediate reasoning step, which the larger model then accepts or rejects before the next step is produced. This step-by-step accept/reject loop requires sequential generation and verification for each reasoning steps. To the best of our knowledge, we are the first work that enables parallel verification for semantic speculative generation.

\section{\sys Design}

Speculative decoding wants two properties that have so far been at odds: \emph{parallel verification}, so the target model is not a sequential bottleneck, and \emph{long acceptance units}, so that each speculation round amortizes its drafting and verification overhead. Token-level methods~\citep{eagle3} verify many drafted tokens in a single target-model pass (parallel), but a single mismatch invalidates the rest of the speculation window, leaving the average accepted span short. Semantic-level methods~\citep{specreason} accept much longer spans by checking whole segments, but they must check one segment, wait for the verdict, then proceed to the next segment (sequential), reintroducing exactly the serial bottleneck speculation was meant to remove. Either way, end-to-end speedups remain bounded.

\sys (PArallel pRefix Speculative Engine) closes this gap. Its core primitive, \emph{parallel prefix verification}, lets a single prefill of the target model emit a Correct/Incorrect decision for \emph{every} prefix of a drafted answer simultaneously. Verification cost is one forward pass regardless of how many prefixes we inspect, so we can recover the long acceptance lengths of semantic-level methods without their sequential verification overhead. At a high level, \sys runs three stages: a small \emph{draft} model writes a complete candidate answer, the target model \emph{holistically verifies} it, and whatever is not accepted is \emph{rewritten} by the target model starting from the longest correct prefix.

\subsection{The \sys Inference Framework}


\sys is built on three empirical observations about how a target model evaluates correctness. To probe this, we hand the target model a draft answer and ask it to emit \texttt{Correct} or \texttt{Incorrect}, then measure how reliably its judgment tracks the actual correctness of the draft. Three findings emerge.

\begin{figure}[h]
  \centering
  \begin{subfigure}[t]{0.33\textwidth}
    \centering
    \includegraphics[width=\linewidth]{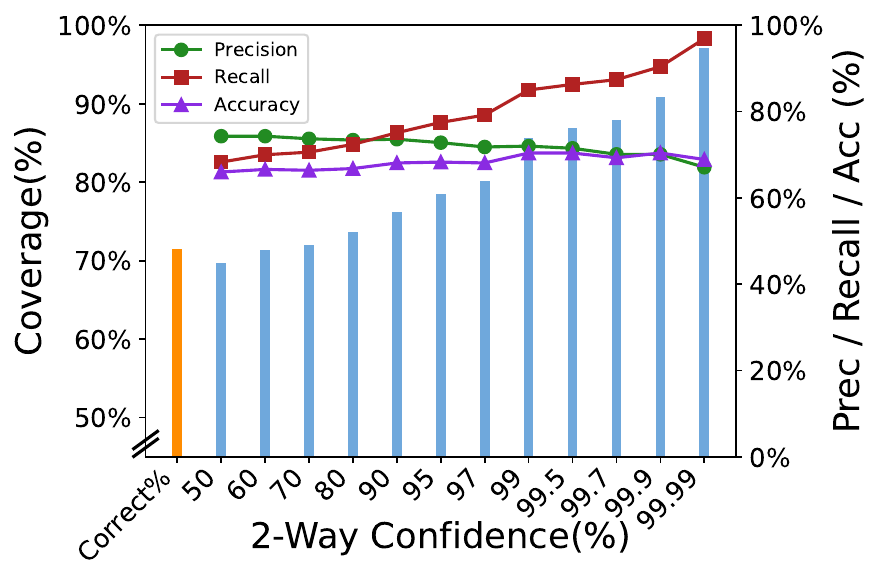}
    \caption{MMLU-Pro}
  \end{subfigure}\hfill
  \begin{subfigure}[t]{0.33\textwidth}
    \centering
    \includegraphics[width=\linewidth]{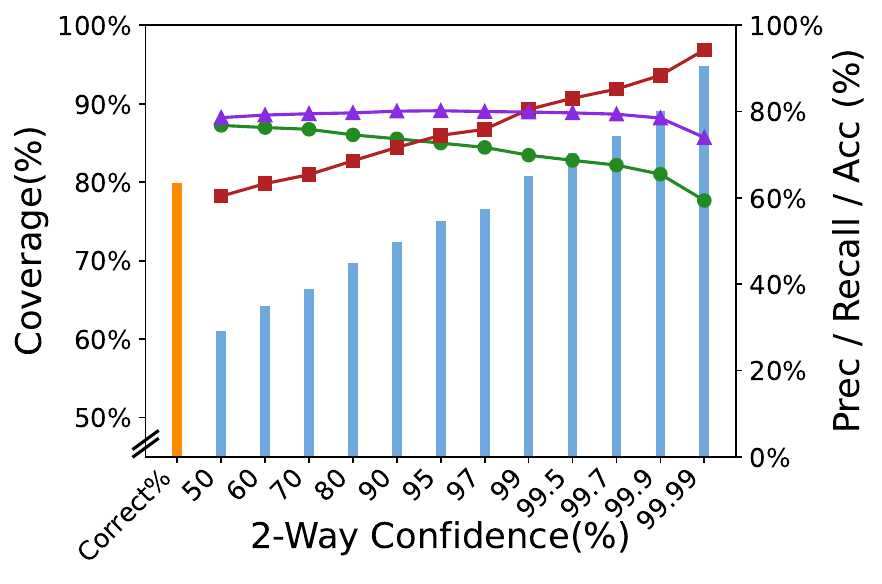}
    \caption{MMLU}
  \end{subfigure}\hfill
  \begin{subfigure}[t]{0.33\textwidth}
    \centering
    \includegraphics[width=\linewidth]{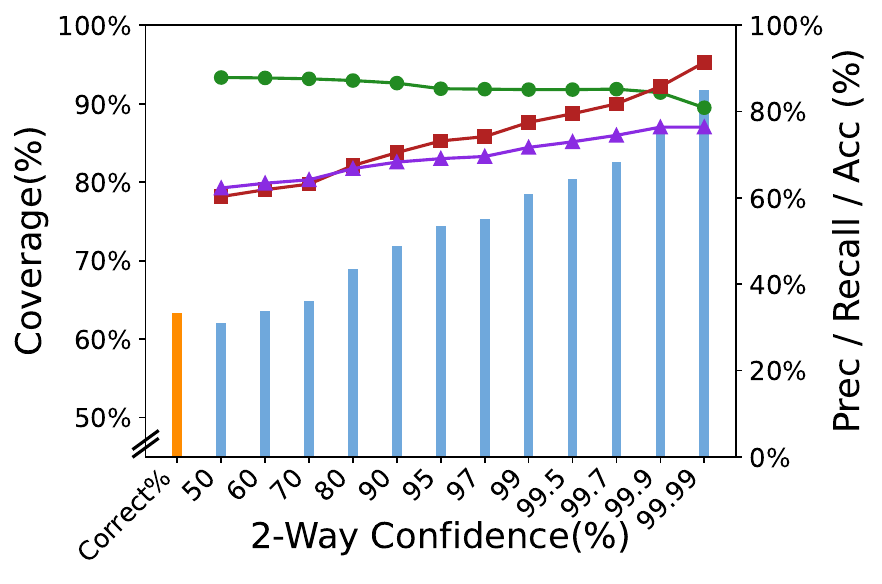}
    \caption{SuperGPQA}
  \end{subfigure}
  \caption{Error-detection metrics on three benchmarks (MMLU-Pro, MMLU, SuperGPQA), with Qwen3-235B prompted to classify a draft answer as \texttt{Correct} or \texttt{Incorrect}. The x-axis is the two-way confidence threshold $\tau$ on $\mathrm{conf} = P(\texttt{Correct})/(P(\texttt{Correct})+P(\texttt{Incorrect}))$; any draft with $\mathrm{conf}<\tau$ is flagged \texttt{Incorrect}. Treating \texttt{Incorrect} as the positive class, the \emph{blue bars} report \emph{coverage} --- the fraction of all drafts the judge flags at threshold $\tau$ (i.e., the rejection rate, including both true and false positives) --- and the line markers report \emph{recall} (of all wrong drafts, the fraction flagged), \emph{precision} (of flagged drafts, the fraction actually wrong), and overall classification \emph{accuracy} (of all drafts, the fraction classified correctly). The \emph{orange bar} separately reports the target model's own accuracy when restricted to the draft's failure cases: among the questions the draft got wrong, how often the target model could itself produce the correct answer.}
  \label{fig:confidence}
\end{figure}

\textbf{Judging is easier than answering.} Even on questions the target model would itself get wrong (orange bars in Fig.~\ref{fig:confidence}), it correctly flags the draft as \texttt{Incorrect} at substantially higher coverage than its own answering accuracy: the model recognizes that something is wrong without knowing the right fix. \sys exploits this asymmetry --- a judge that detects errors well, even when it cannot produce the right answer itself, is enough to supervise a fast, error-prone draft model.


\textbf{Confidence beats argmax.} The first principle only pays off if we can extract the judge's verdict reliably, and the naive readout — taking whichever of \texttt{Correct} or \texttt{Incorrect} the model assigns higher probability — does not: this argmax decision catches just  $60$--$70\%$ of the draft's errors (Fig.~\ref{fig:confidence}, $\tau=0.5$). The probabilities themselves carry the model's uncertainty, but collapsing them to a single token discards it. Defining the \emph{two-way confidence}
\[
\mathrm{conf}(y) \;=\; \frac{P(\texttt{Correct})}{P(\texttt{Correct})+P(\texttt{Incorrect})},
\]
and treating any draft with $\mathrm{conf} < \tau$ as suspect, we catch nearly all errors --- the recall (fraction of wrong drafts flagged) jumps from $\sim$$65\%$ at the argmax to nearly $100\%$ at $\tau=0.997$.

\textbf{Aggressive error-catching does not waste good drafts.} A higher $\tau$ catches more errors, but in principle it could also flag many \emph{correct} drafts as wrong --- and every such false alarm forces an unnecessary target-model regeneration. Fig.~\ref{fig:confidence} shows this does not happen: as \emph{recall} (of all wrong drafts, the fraction we flag) climbs with $\tau$, both \emph{precision} (of the drafts we flag, the fraction that are actually wrong) and overall classification \emph{accuracy} (of \emph{all} drafts, the fraction we classify correctly) stay flat. Concretely, the confidence margin separates correct from incorrect drafts cleanly enough that almost every correct draft scores well above $\tau=0.997$, so raising the threshold captures additional errors without misclassifying correct drafts. \sys can therefore afford to set $\tau$ aggressively.

These observations leads to the design of \sys, which consists of 3 stages: a small draft model proposes a complete answer, the confidence-thresholded judge (the target model) decides what to keep, and the target model fills in whatever the judge rejects (Fig.~\ref{fig:pipeline_example}).

\noindent
\begin{minipage}[t]{0.40\linewidth}
\begin{algorithm}[H]
\tiny
\caption{\sys: draft $\to$ holistic verify $\to$ rewrite}
\label{alg:draft-rescue}
\DontPrintSemicolon
\textbf{Parameters:} confidence threshold $\tau$; chunk size $\Delta$.\;
\ForEach{question $q$}{
  \gcomment{Stage 1: draft generation}
  $y_s \gets \textsc{SmallDraft}(q)$\;
  \gcomment{Stage 2(i): full-answer judgment}
  $(v,\,\mathrm{conf}) \gets \textsc{JudgeFull}(q, y_s)$\;
  \uIf{$v=\texttt{Correct}$ \textbf{and} $\mathrm{conf}\ge\tau$}{
     \gcomment{Stage 3: accept}
     \Return $y_s$\;
  }
  \Else{
     \gcomment{Stage 2(ii): parallel prefix verification (\S\ref{sec:ppv})}
     $t^\star \gets \textsc{VerifyPartial}(q,\, y_s;\, \Delta)$\;
     \uIf{$t^\star \neq \varnothing$}{
        \gcomment{Stage 3: continue}
        $y_{\text{cont}} \gets \textsc{BigContinue}(q,\ y_{1:t^\star})$\;
        \Return $y_{1:t^\star} \,\Vert\, y_{\text{cont}}$\;
     }
     \Else{
        \gcomment{Stage 3: restart}
        \Return $\textsc{BigFromScratch}(q)$\;
     }
  }
}
\end{algorithm}
\end{minipage}\hfill
\begin{minipage}[t]{0.58\linewidth}
\centering
\vspace*{0pt}
\includegraphics[width=\linewidth]{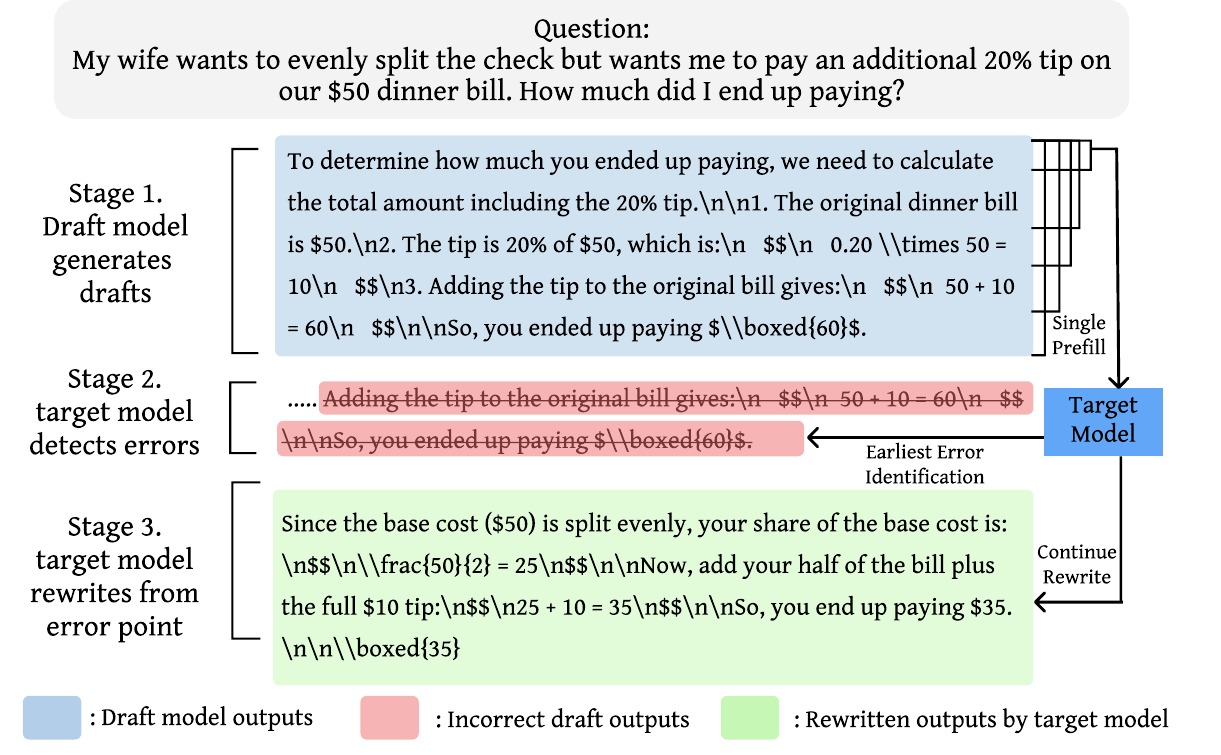}
\captionof{figure}{A concrete run of \sys. The draft model errs starting from the fourth line (strikethrough red text) and outputs 60. The target model locates the error and regenerates the suffix, producing the correct answer 35.}
\label{fig:pipeline_example}
\end{minipage}

\paragraph{Stage 1 --- a cheap candidate.}
The draft model produces a full candidate answer $y_{1:T}$ for the prompt $q$. We use \texttt{Qwen3-8B}, which under SGLang~\citep{zheng2024sglang} decodes at roughly $4\times$ the speed of our Qwen3 235B taget model. An acceptance rate of $\sim$$25\%$ on draft tokens therefore already matches baseline latency theoretically, and anything above is net speedup. \sys is generous about what it asks of the draft: it only needs the draft to be \emph{adequate} on average, since the Stage~2 judge catches the rest. Modern open small models comfortably reach this bar and typically do far better.

\paragraph{Stage 2 --- verifying the draft.}
We apply the confidence-thresholded judge to the candidate twice, and both passes are prefill-only and KV-cache-light, so they stay cheap even on a large target model. \emph{First, on the full draft.} A single prefill of $(q, y_{1:T})$ yields $\mathrm{conf}(y_{1:T})$; if it reaches $\tau$, we accept and stop. \emph{Second, on prefixes.} If the full draft fails, we want the longest prefix the same target model would still accept---the largest $t^\star$ with $\mathrm{conf}(y_{1:t^\star}) \ge \tau$. Most draft errors are localized to a span late in the answer, so the maximum-prefix rule reuses every token before the first error and avoids paying to regenerate work that was already correct. Naively this would mean one prefill per candidate boundary; we describe in \S\ref{sec:ppv} a custom attention-mask construction that collapses all $n$ prefix judgments into a single prefill, keeping this pass prefill-only as well.

\paragraph{Stage 3 --- acting on the judgement.}
The judge's outcome selects one of three branches (Alg.~\ref{alg:draft-rescue}). If $\mathrm{conf}(y_{1:T}) \ge \tau$, we \textbf{accept} the full draft. If a longest accepted prefix $y_{1:t^\star}$ was found, we \textbf{continue}: the target model resumes generation from position $t^\star{+}1$, reusing the draft up through $t^\star$. Otherwise --- no prefix's confidence reaches $\tau$, or every prefix passes leaving no clear error boundary --- we \textbf{restart} from scratch with the target model.

\paragraph{Compatibility with token-level speculative decoding.}
\sys operates purely at the semantic level and is orthogonal to token-level speculation: methods such as EAGLE3~\citep{eagle3} plug cleanly into Stage~1 (accelerating the draft worker) and into the target-model continuation/regeneration of Stage~3, without touching \sys's verification logic. The two mechanisms shorten different parts of the decoding budget. \sys shortens the \emph{length} of the sequence the target must generate, by reusing whichever draft prefix survives verification and resuming only from the first error. Token-level speculation shortens the \emph{time per generated token}, by letting the target verify several drafted next-tokens in a single forward pass. Because one mechanism reduces the number of tokens decoded and the other reduces the latency of each remaining token, their savings compose roughly multiplicatively; Table~\ref{tab:hover-results} (row \sys{}+E3) and Figure~\ref{fig:combine} quantify this composition.

\begin{figure}[h!]
    \centering
    \includegraphics[width=\linewidth]{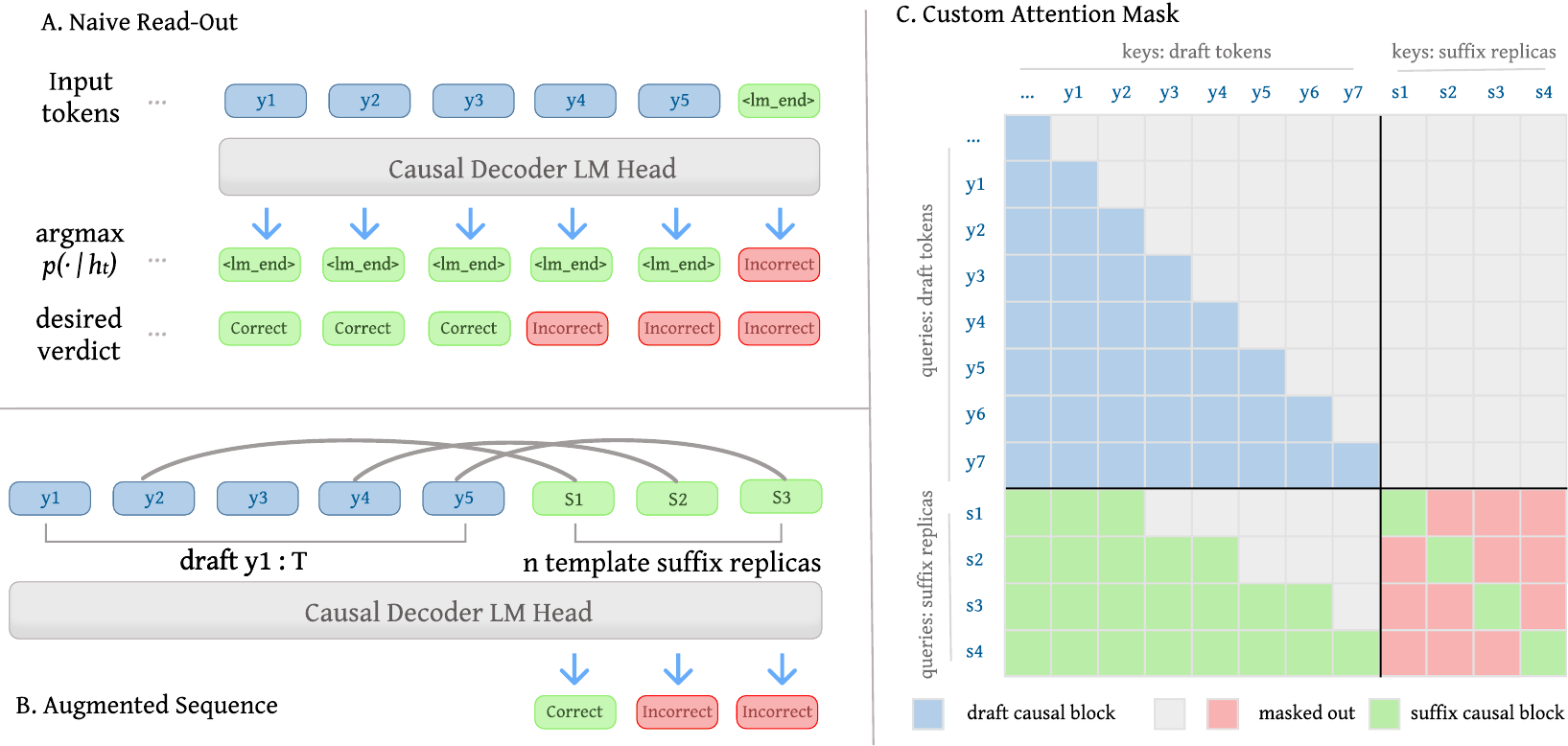}
    \caption{Parallel prefix verification via augmented chat-template suffixes. Naive token-dimension segmentation makes the model emit chat-template tokens instead of Correct/Incorrect; appending a duplicated template suffix per prefix, isolated by a custom attention mask, elicits $n$ independent Correct/Incorrect classifications in one pass.}
    \label{fig:smart-segment}
\end{figure}

\subsection{Parallel Prefix Verification}
\label{sec:ppv}

Given a draft $y_{1:T}$ rejected in Stage~2(i), we need the largest $t^\star$ such that $y_{1:t^\star}$ would pass the same Correct/Incorrect test. The naive implementation runs one target-model prefill per candidate boundary, at a cost linear in the number of boundaries $n$ --- several extra prefills of the large target model per question, which is enough to erase the speedup the draft was meant to provide.

\sys utilizes a simple fact about prefill: \emph{a causal decoder already computes a hidden state at every position of $y_{1:T}$}. A single prefill of $(q, y_{1:T})$ produces $T$ hidden states --- one per token --- each of which summarizes the prefix up to that point and could in principle drive a Correct/Incorrect classification at that boundary, if we knew how to read it out. The classification is read out by prompting the target to emit a judgment, similar to the Stage~2 full-answer pass. So if we can extract such a judgment at each of the $n$ candidate boundaries directly from these already-computed hidden states, we verify all $n$ prefixes for the cost of roughly one prefill instead of $n$.

The obstacle is a \emph{format mismatch}. The target model is an instruction-tuned chat model, trained to emit a Correct/Incorrect token \emph{only when the input ends in the chat-template suffix that requests that classification} (e.g., \texttt{...<|im\_end|><|im\_start|>assistant}). At an arbitrary boundary $t_i$ in the middle of a draft, that suffix has not been seen --- the input ends mid-utterance, inside an assistant turn that the model believes is still in progress. Decoding from the hidden state at $t_i$ therefore produces the chat-template tokens the model expects to emit next (turn boundaries, role markers, end-of-message continuations), not a Correct/Incorrect verdict (Fig.~\ref{fig:smart-segment}, left). The hidden state has the right \emph{information} about the prefix, but the model's decoding policy refuses to emit it unless we put the right scaffolding.

We resolve this with a custom attention mask construction (Fig.~\ref{fig:smart-segment}). We keep the draft tokens $y_{1:T}$ as is and append $n$ \emph{copies} of the chat-template suffix, one per candidate boundary, into the same sequence. A custom attention mask wires up visibility cleanly: the draft tokens attend causally among themselves, so a single KV cache for $y_{1:T}$ is computed once and shared across all $n$ classification queries; each suffix copy $i$ attends causally to the draft tokens up through its assigned boundary $y_{1:t_i}$ and to nothing else --- in particular, suffix copies do not see each other. The classification position at the end of suffix $i$ therefore sees \emph{exactly} the input it would have seen had we run the single $(y_{1:t_i}, \text{suffix})$ pair on its own; the $n$ readouts are independent by construction. One forward pass yields $n$ Correct/Incorrect predictions, from which we read off $t^\star = \max\{t_i \mid y_{1:t_i}\ \text{is Correct}\}$ (\textsc{VerifyPartial} in Algorithm~\ref{alg:draft-rescue}). The total cost is one prefill of length $T + n\cdot|\text{suffix}|$ --- a small additive overhead over a single full-answer prefill --- regardless of $n$. We place the boundaries uniformly every $\Delta$ tokens (so $n \approx T/\Delta$), trading verification granularity against per-suffix overhead.

\section{Experiments}

\subsection{Setup}
We evaluate \sys in two configurations. The \emph{primary} configuration uses Qwen3-235B-A22B-Instruct-FP8~\citep{qwen3} as the target model and Qwen3-8B as the draft. To test generalization beyond the Qwen models, and whether our confidence-based judgement holds beyond the Qwen models, we also run a \emph{cross-family} configuration with GLM-4.7-FP8~\citep{glm} as the target and Qwen3.5-9B~\citep{qwen3.5} as the draft (different pretraining data, tokenizer, and chat template). To verify that \sys composes with token-level speculative decoding, we additionally integrate EAGLE3~\citep{eagle3} on both the draft and target workers (\sys+E3); using EAGLE3 alone (\emph{Eagle3}) on the target is reported as a reference. We implement \sys on top of the SGLang inference engine~\citep{zheng2024sglang}, with no retraining of either the draft or the target model.

We also compare against SpecReason~\citep{specreason}, a semantic-level speculative generation method that verifies reasoning steps sequentially. We do not compare against Judge Decoding~\citep{judge_decoding}, as its training requires undisclosed dataset, and it can be viewed as an improved token-level speculative decoding method that could be composed with \sys, similar to EAGLE3.

We evaluate on four task categories: \textbf{General} (MMLU-Redux~\citep{mmluredux}, MMLU-Pro~\citep{mmlupro}, GPQA~\citep{gpqa}), \textbf{Math \& STEM} (MATH~\citep{MATH}, GSM8K~\citep{gsm8k}), \textbf{Coding} (HumanEval~\citep{humaneval}, MBPP~\citep{MBPP}), and \textbf{Open-ended chat} (MT-Bench~\citep{mtbench}, multi-turn questions scored $1$--$10$ by Gemini 3 Flash). We use the model's instruct mode with no CoT or reasoning enabled. For multi-choice datasets (MMLU, GPQA), the model is prompted to give explanation before its final answer rather than giving a single letter. All runs use greedy decoding on 4 \texttt{NVIDIA H200 GPUs}.

\subsection{Accuracy and Speedup}

\begin{table}[t]
\centering
\renewcommand{\arraystretch}{1.15}
\setlength{\tabcolsep}{6pt}
\footnotesize

\caption{\textbf{\sys vs.\ Baselines} on Qwen3-235B-A22B (FP8, SGLang).
235B = plain autoregressive; 8B = Qwen3-8B standalone (draft model);
Eagle3 = speculative decoding on 235B;
\sys = 8B draft with full and partial verify (no Eagle3);
+E3 = \sys with EAGLE3 on both draft and target.}
\label{tab:hover-results}

\vspace{6pt}

\begin{tabular}{l ccc cccc ccc}
\toprule
\rowcolor{headerbg}
\textcolor{headertext}{\textbf{Benchmark}} &
\multicolumn{3}{c}{\textcolor{headertext}{\textbf{Accuracy / Score}}} &
\multicolumn{4}{c}{\textcolor{headertext}{\textbf{TPS (tok/s)}}} &
\multicolumn{3}{c}{\textcolor{headertext}{\textbf{Speedup vs.\ 235B}}} \\

\cmidrule(lr){2-4} \cmidrule(lr){5-8} \cmidrule(lr){9-11}

& 8B & 235B & \sys & 235B & Eagle3 & \sys & +E3 & Eagle3 & \sys & +E3 \\
\midrule

\rowcolor{lightgray}
MMLU
  & 0.776 & 0.868 & 0.828
  & 89.5 & 145.2 & 130.4 & 216.6
  & \speedup{1.62} & \speedup{1.46} & \speedup{2.42} \\

MMLU-Pro
  & 0.568 & 0.668 & 0.636
  & 92.3 & 137.0 & 149.5 & 191.3
  & \speedup{1.48} & \speedup{1.62} & \speedup{2.07} \\

\rowcolor{lightgray}
GPQA
  & 0.364 & 0.460 & 0.424
  & 95.3 & 129.7 & 118.7 & 151.7
  & \speedup{1.36} & \speedup{1.25} & \speedup{1.59} \\

MATH
  & 0.764 & 0.708 & 0.756
  & 96.1 & 164.3 & 215.7 & 272.2
  & \speedup{1.71} & \speedup{2.24} & \speedup{2.83} \\

\rowcolor{lightgray}
GSM8K
  & 0.940 & 0.948 & 0.952
  & 94.3 & 176.4 & 405.0 & 420.7
  & \speedup{1.87} & \speedup{4.29} & \speedup{4.46} \\

MT-Bench
  & 7.93 & 9.19 & 8.88
  & 95.6 & 153.9 & 167.5 & 194.0
  & \speedup{1.61} & \speedup{1.75} & \speedup{2.03} \\

\midrule

\rowcolor{lightgray}
MBPP+
  & 0.669 & 0.799 & 0.746
  & 86.9 & 152.6 & 148.1 & 170.9
  & \speedup{1.76} & \speedup{1.70} & \speedup{1.97} \\

HumanEval+
  & 0.707 & 0.835 & 0.872
  & 92.6 & 166.6 & 149.6 & 193.1
  & \speedup{1.80} & \speedup{1.62} & \speedup{2.08} \\

\bottomrule
\end{tabular}

\end{table}

Table~\ref{tab:hover-results} reports accuracy and throughput on the primary Qwen3-235B setup, comparing the 235B target alone, the 8B draft alone, EAGLE3~\citep{eagle3} on the target, \sys, and \sys{}+E3.

\textbf{Accuracy.} \sys's accuracy sits between the draft and the target on every benchmark, tracking the 235B target within a few points and lifting the draft by a wide margin. The closeness to the target's accuracy across reasoning, knowledge, and coding tasks indicates that the improvement comes from the framework's error-identification mechanism rather than the draft's own capacity.

\textbf{Speedup.} \sys delivers consistent speedups over the 235B target across all benchmarks. Composing \sys with EAGLE3 (\sys{}+E3) yields roughly multiplicative gains, because the two mechanisms attack complementary bottlenecks: \sys reduces how many tokens the target has to decode at all, while EAGLE3 accelerates each token the target does decode. The size of the speedup varies with task difficulty. On easier tasks (GSM8K, MATH), the draft is correct most of the time, so \sys frequently accepts the full draft and the gains are largest. On harder tasks (GPQA, MMLU-Pro), the draft is wrong more often and \sys has to fall back to prefix verification or full regeneration; the lower acceptance rate translates into a smaller speedup.

\subsection{Comparison with SpecReason}
We compare with SpecReason~\citep{specreason}, a representative semantic-level speculative generation method, by adapting its code for general text generation instead of chain-of-thought-only reasoning. SpecReason generates text and, at fixed checkpoints, asks the large model for a $1$--$10$ quality score to decide whether to accept the latest segment or have the large model rewrite it. For a fair comparison, we set SpecReason's chunk size to match \sys's verification boundaries, as configured in Appendix~\ref{appendix:specreason-config}.

Figure~\ref{fig:pareto-sr-vs-hover} shows accuracy against throughput for both methods. \sys reaches accuracy on par with SpecReason on every benchmark, and at substantially higher throughput. The throughput advantage comes from \emph{parallel} verification: SpecReason judges segments sequentially at fixed checkpoints, so each checkpoint must wait for the previous one's verdict before generation can proceed, while \sys verifies all prefixes of the draft in a single forward pass and reads off the longest accepted boundary.

\begin{figure}[!htb]
\centering
\begin{minipage}[c]{0.42\linewidth}
  \centering
  \includegraphics[width=\linewidth]{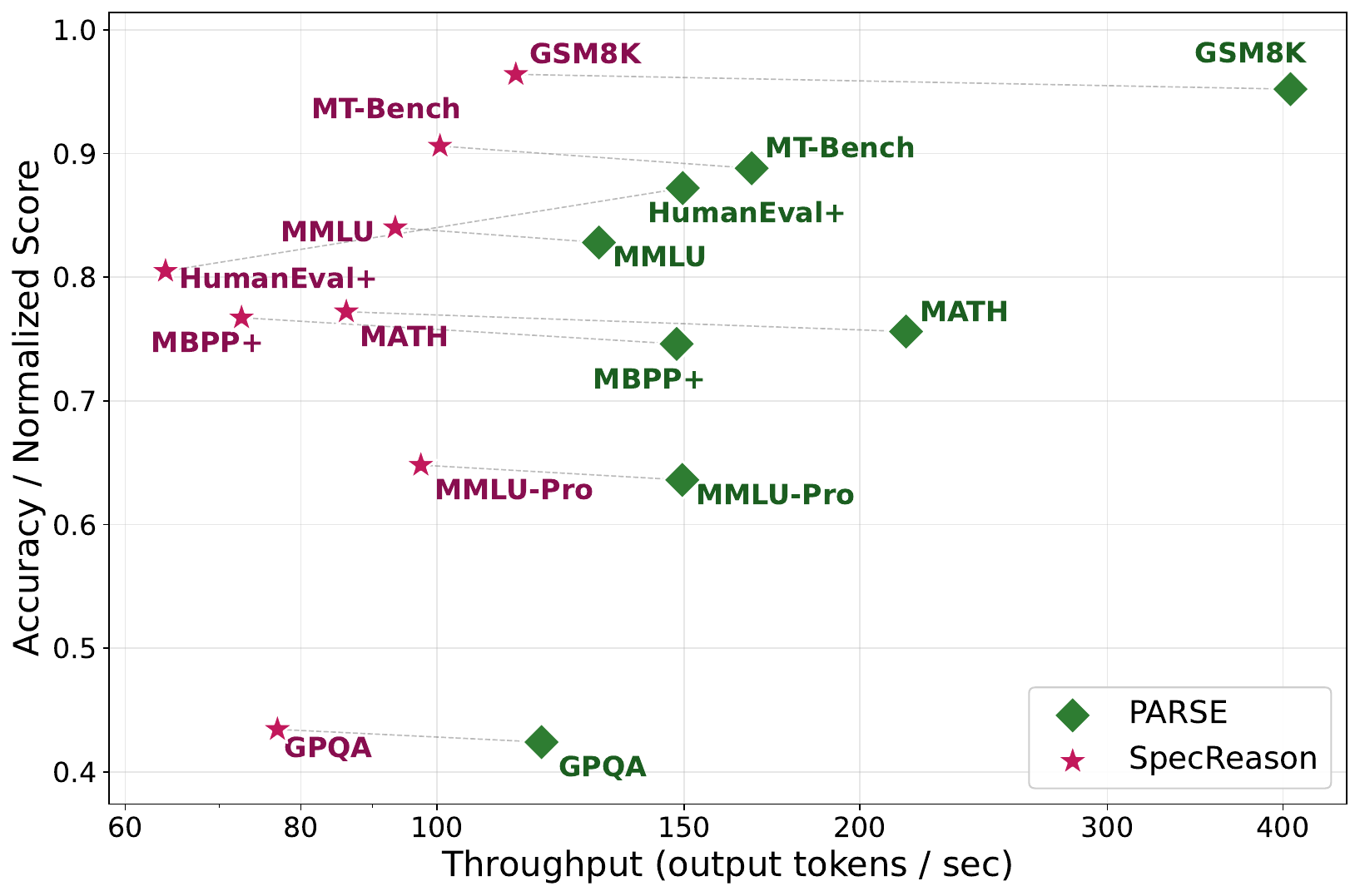}
  \caption{Accuracy vs.\ throughput of \sys and SpecReason on Qwen3-235B. Higher and farther right is better; dashed lines connect the same benchmark across methods.}
  \label{fig:pareto-sr-vs-hover}
\end{minipage}\hfill
\begin{minipage}[c]{0.56\linewidth}
  \centering
  \renewcommand{\arraystretch}{1.0}
  \setlength{\tabcolsep}{2pt}
  \scriptsize
  \begin{tabular}{l ccc cc c}
    \toprule
    \rowcolor{headerbg}
    \textcolor{headertext}{\textbf{Benchmark}} &
    \multicolumn{3}{c}{\textcolor{headertext}{\textbf{Accuracy / Score}}} &
    \multicolumn{2}{c}{\textcolor{headertext}{\textbf{TPS (tok/s)}}} &
    \textcolor{headertext}{\textbf{Speedup}} \\
    \cmidrule(lr){2-4} \cmidrule(lr){5-6} \cmidrule(lr){7-7}
    & GLM-4.7 & Qwen3.5 & \sys & GLM-4.7 & \sys & \\
    \midrule
    \rowcolor{lightgray}
    MMLU       & 0.860 & 0.780 & 0.860 & 70.0 & 104.2 & \speedup{1.49} \\
    MMLU-Pro   & 0.720 & 0.636 & 0.712 & 70.0 &  78.2 & \speedup{1.12} \\
    \rowcolor{lightgray}
    GPQA       & 0.596 & 0.434 & 0.581 & 70.3 &  66.8 & \speeddown{0.95} \\
    MATH       & 0.796 & 0.728 & 0.784 & 70.2 & 126.0 & \speedup{1.79} \\
    \rowcolor{lightgray}
    GSM8K      & 0.956 & 0.916 & 0.940 & 69.4 & 175.7 & \speedup{2.53} \\
    MT-Bench$^{\dagger}$ & 9.08 & 8.85 & 9.11 & 69.2 & 85.3 & \speedup{1.23} \\
    \midrule
    \rowcolor{lightgray}
    MBPP+      & 0.780 & 0.693 & 0.738 & 65.3 & 110.8 & \speedup{1.70} \\
    HumanEval+ & 0.805 & 0.732 & 0.805 & 66.4 & 147.0 & \speedup{2.21} \\
    \bottomrule
  \end{tabular}

  \vspace{4pt}

  \captionof{table}{Cross-family generalization: \sys with GLM-4.7-FP8 as the target model and Qwen3.5-9B as the cross-family draft, vs.\ GLM-4.7-FP8 and Qwen3.5-9B standalone baselines.}
  \label{tab:hover-glm-results}
\end{minipage}
\end{figure}

\subsection{Cross-model Generalization}
To validate that \sys's error-identification mechanism does not depend on the draft and target sharing a pretraining lineage, we evaluate on a deliberately cross-family pairing: \texttt{GLM-4.7-FP8} as the target model and \texttt{Qwen3.5-9B} as the draft, with different tokenizers, pretraining data, and chat templates. Results are in Table~\ref{tab:hover-glm-results}.

Accuracy tracks the GLM-4.7 baseline closely across all benchmarks. \sys delivers consistent speedups on most benchmarks. The limitation is GPQA, where the cross-family pairing dips slightly below the baseline: the draft and the target disagree often enough that partial-verification overhead outweighs the benefit of draft reuse. Overall, this confirms that the confidence-based error identification generalizes across model families, while the magnitude of the speedup still depends on how often the draft produces prefixes the target is willing to accept.



\subsection{Acceleration Breakdown}

\begin{figure}[h]
    \centering
    \begin{subfigure}[t]{0.47\linewidth}
        \centering
        \includegraphics[width=\linewidth]{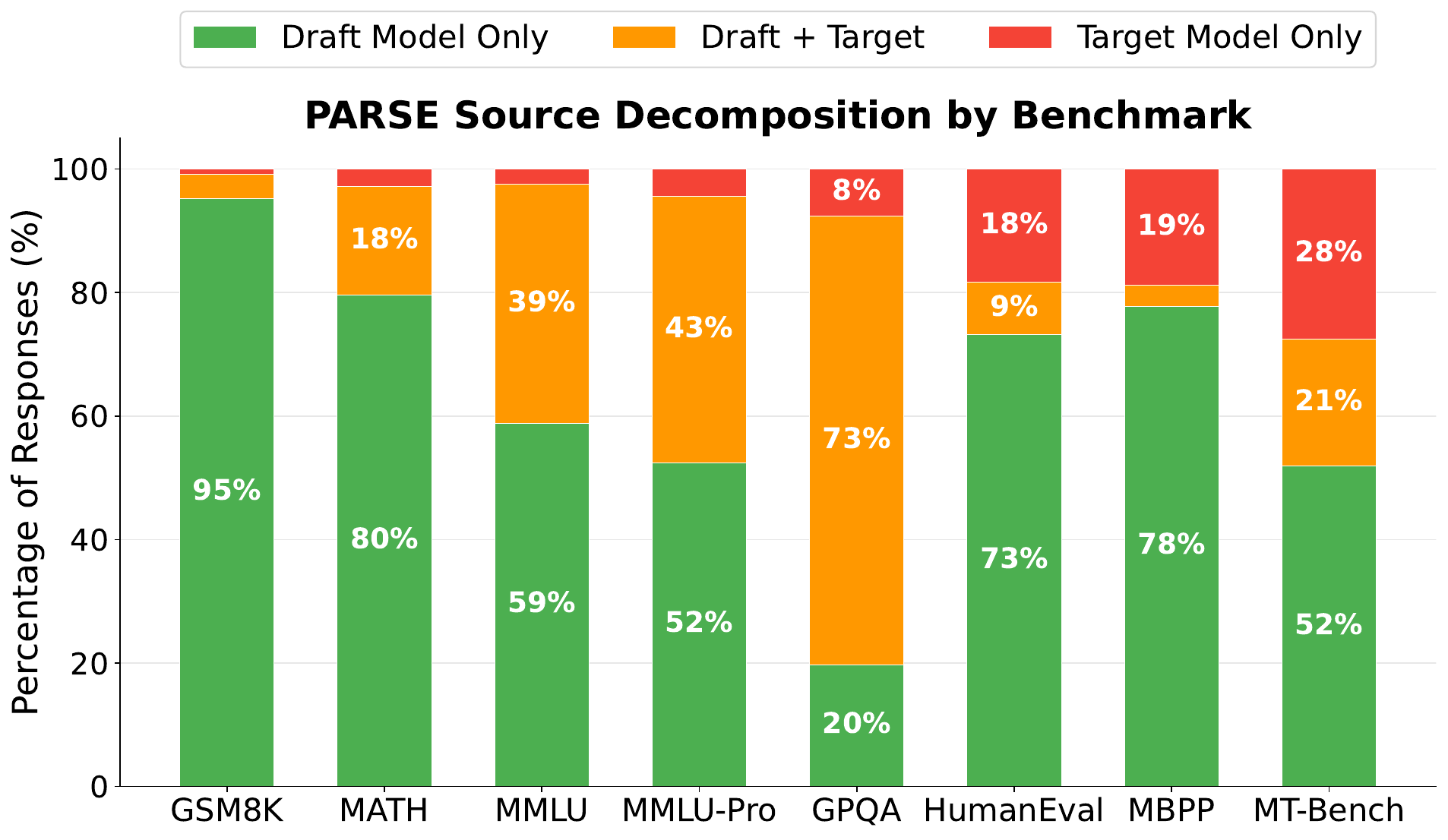}
        \caption{Source breakdown of final responses. \textit{Draft}: draft accepted; \textit{Target}: regenerated from scratch with target model; \textit{Mixed}: continued from a partial draft.}
        \label{fig:stackbar}
    \end{subfigure}
    \hfill
    \begin{subfigure}[t]{0.515\linewidth}
        \centering
        \includegraphics[width=\linewidth]{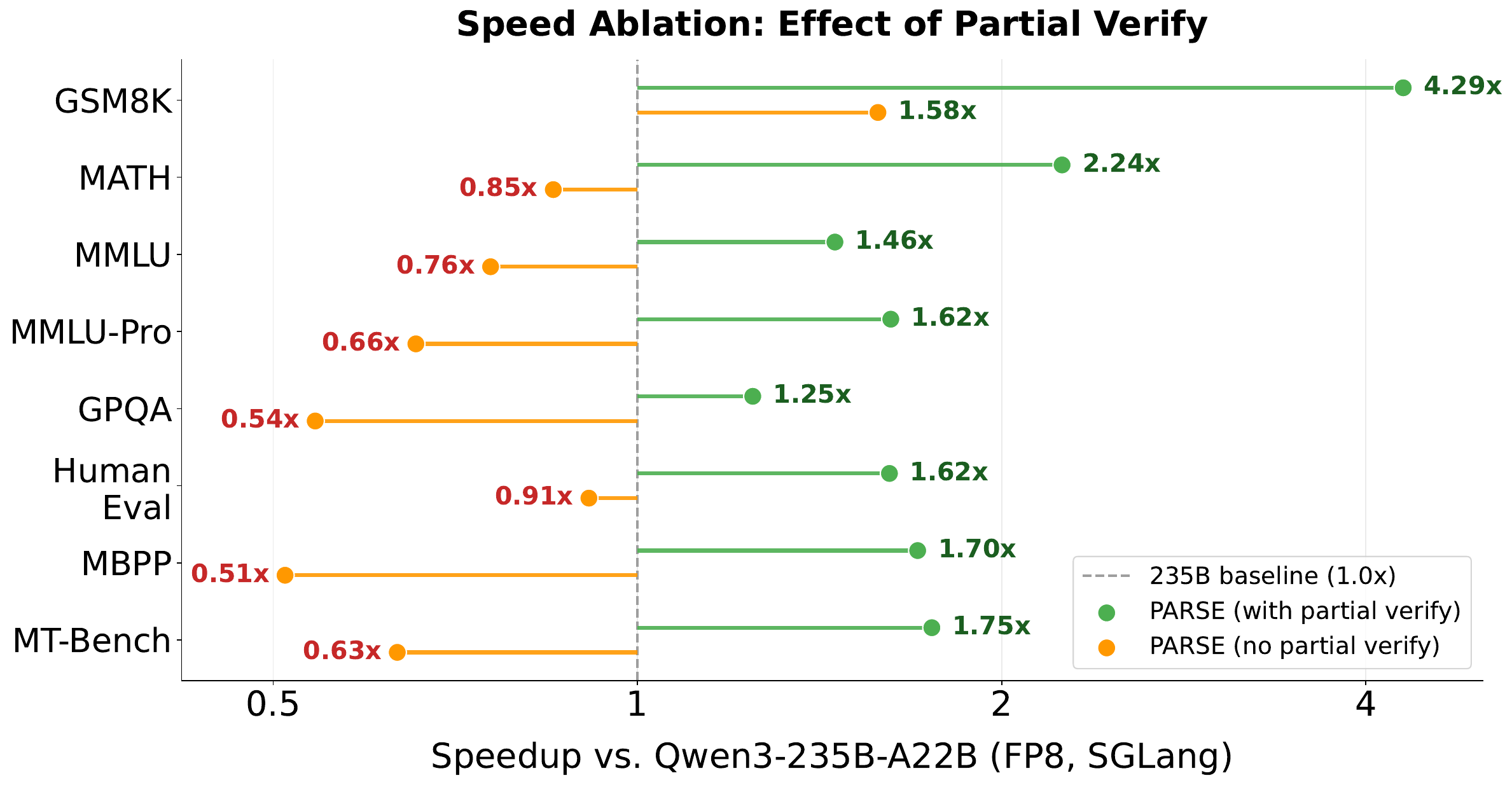}
        \caption{Speed with Qwen3-235B Target Model only, \sys without partial verification, and \sys with partial verification}
        \label{fig:speed}
    \end{subfigure}
\end{figure}

Fig.~\ref{fig:stackbar} breaks down the sources of final responses, showing the proportion of answers that come from drafts directly accepted, full regeneration by the target model, or continuation from a partially accepted draft. The distribution varies substantially across datasets. On easier benchmarks such as GSM8K, a large fraction of responses are accepted directly from the draft model, yielding significant speedups. On more challenging datasets such as GPQA, fewer drafts can be accepted outright; however, \sys effectively reuses partial drafts by continuing from the last correct prefix, enabling faster regeneration compared to restarting from scratch.

\subsection{Ablation Experiment}

\begin{figure}[h]
    \centering
    \begin{minipage}{0.49\linewidth}
        \centering
        \includegraphics[width=\linewidth]{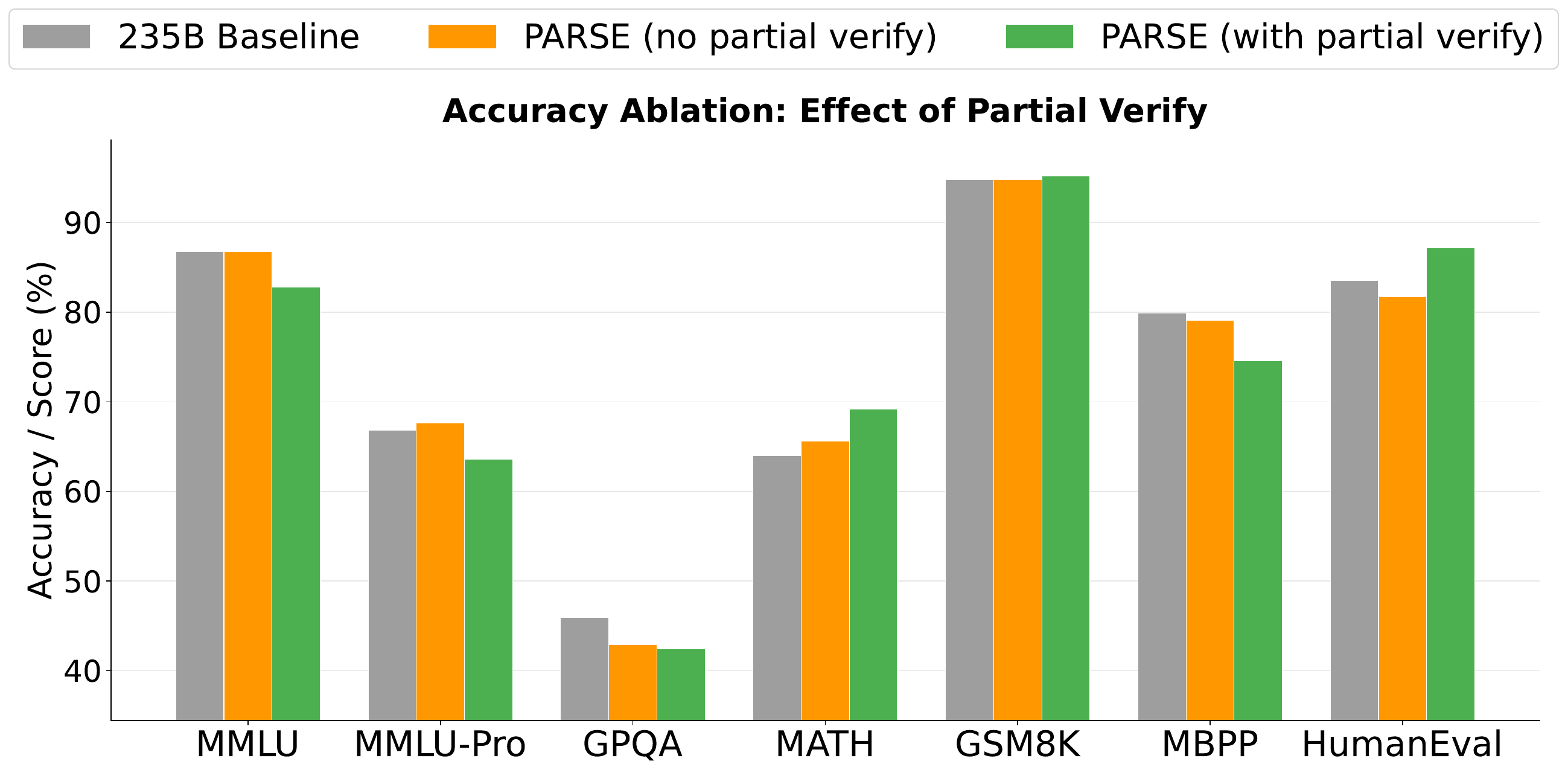}
        \caption{Accuracy with Qwen3-235B, \sys without partial verification, and \sys with partial verification.}
        \label{fig:ablation_acc}
    \end{minipage}
    \hfill
    \begin{minipage}{0.48\linewidth}
        \centering
        \includegraphics[width=\linewidth]{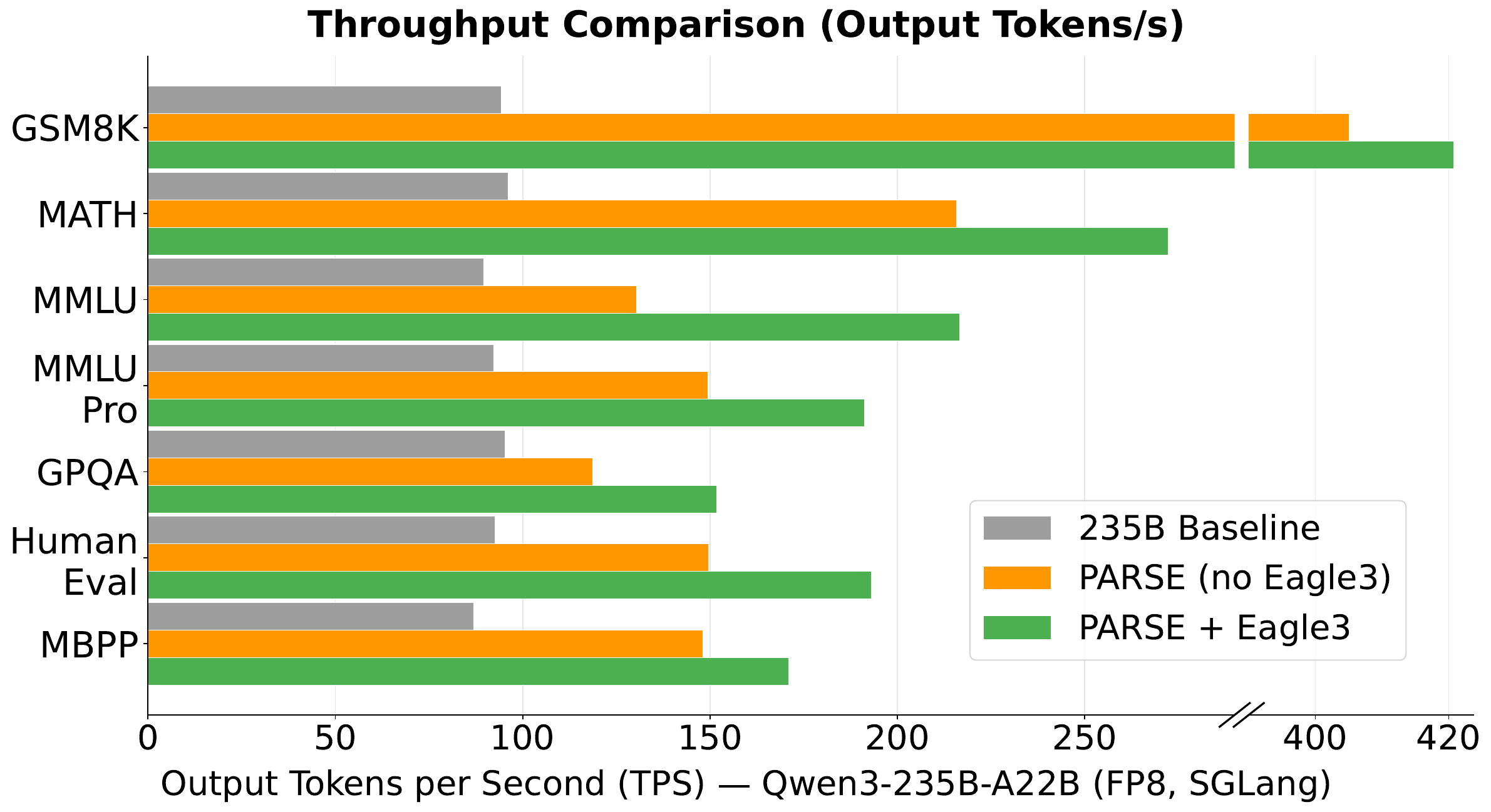}
        \caption{Throughput across datasets for the Qwen3-235B baseline, \sys{}, and \sys{} combined with EAGLE3 on both workers.}
        \label{fig:combine}
    \end{minipage}
\end{figure}

We conduct an ablation study to isolate the contribution of partial prefix verification. The ablation compares three settings: the Qwen3-235B target alone, \sys without partial verification (a rejected full-answer judgment forces the target to restart from scratch), and full \sys with partial verification (the target resumes from the longest accepted prefix).

\textbf{Speed.} Fig.~\ref{fig:speed} shows that partial verification is the dominant driver of \sys's speedup. Without it, every rejected draft costs a full target-model regeneration, and \sys runs \emph{slower} than the target baseline on most benchmarks. With partial verification, \sys reuses the correct portion of every rejected draft and recovers the full speedup of Table~\ref{tab:hover-results}.

\textbf{Accuracy.} Fig.~\ref{fig:ablation_acc} reports accuracy for the same three settings. \sys with partial verification matches the 235B baseline within a small margin on every benchmark. \sys without partial verification is similarly close to the baseline on accuracy, confirming that the speed advantage of partial verification comes without sacrificing too much generation quality.

\textbf{Combination with token-level speculative decoding.}
\sys is orthogonal to token-level speculative decoding, so the two can be composed for additional speedup. We demonstrate this with EAGLE3~\citep{eagle3} integrated into SGLang: EAGLE3 runs inside the draft worker (Stage~1) and during the target's continuation (Stage~3) without changing \sys's verification logic. Fig.~\ref{fig:combine} reports the per-benchmark throughput of \sys{}+Eagle3 against \sys alone and the 235B baseline. The composition strictly improves over either method alone on every benchmark.

\subsection{Case Studies}
\label{case_study}

We manually inspect 200 cases from the experiment results where the draft model's response is partially correct, and the larger model is able to identify the beginning of the errors and continue from the last correct position. We show one such example here.

In this example, black text shows the shared prefix; red marks the first incorrect span found by the target model in a single pass; green text is the regenerated suffix. \sys then outputs black + green text as the final answer, demonstrating that the target model reliably locates the last correct position and efficiently resumes generation, maximizing token reuse.

\newcommand{\HRule}{\noindent\rule{\linewidth}{1pt}}

\HRule

\noindent\textbf{Question:} Kenny is selling his \textit{Pokemon} cards to buy a ticket to an amusement park, which costs \$100. He has a collection of cards and plans to sell them for \$1.5 each. He keeps 1/3 of them and gets to go to the amusement park with \$50 in spending cash. How many cards did he start with?

\HRule

Let’s break down the problem step by step. Kenny wants to buy a ticket to an amusement park that costs \$100. He has a collection of \textit{Pokemon} cards, and he plans to sell them for \$1.50 each. He keeps 1/3 of the cards, meaning he sells 2/3 of them. After selling the cards, he has \$50 in spending cash. We are to find out how many cards Kenny started with.
\noindent Step 1: Let the total number of cards be $x$. He sells $\frac{2}{3}x$ cards. Each card is sold for \$1.50, so the total money he gets from selling the cards is:
\textit{Money from selling cards} = $1.5 \times \frac{2}{3}x = 1.0x$.
So, he earns \$x from selling \textcolor{red}{the cards. He has \$50 in spending cash, and he needs \$100 for the ticket. That means he needs an additional \$50. So, the money he earned from selling the cards must be \$50: $x = 50$ --- Final Answer: 50. Kenny started with 50 cards.} \textcolor{ForestGreen}{the cards. This money is used to buy a \$100 ticket and leaves him with \$50 in spending cash. So the total money he earns from selling the cards is: \$100 + \$50 = \$150. We already found that the money earned is $x$ dollars (since $1.5 \times \tfrac{2}{3}x = x$). So: $x = 150$. Therefore, Kenny started with 150 \textit{Pokemon} cards.}

\HRule

\section{Conclusion}
\label{sec:conclusion}

We presented \sys, a semantic-level speculative generation framework built on \emph{parallel prefix verification}: one prefill judges every prefix of a draft via duplicated chat-template suffixes and a custom attention mask. On Qwen3-235B(FP8), \sys delivers $1.25\times$--$4.3\times$ end-to-end speedup over the target with target-level accuracy, and $1.6\times$--$4.5\times$ when composed with EAGLE3.

\newpage
\bibliography{references}
\bibliographystyle{references}

\newpage
\appendix


\section{Detailed \sys{} Algorithm}
\label{app:algo-detailed}

This appendix expands the concise main-text Algorithm~\ref{alg:draft-rescue}
into the full procedure that we actually run in the engine, including
every decision branch, every tunable hyperparameter, and the
two-way-confidence calibration of judge logits.

\subsection{Notation and judge calibration}
\label{app:algo-notation}

Let $q$ denote the user prompt, $M_s$ the small \emph{draft} model, and
$M_b$ the large \emph{target} model, which is also reused as the
\emph{verifier} $J$ so that no extra parameters are trained or loaded.
A draft answer $y_s=(y_1,\dots,y_T)$ is produced by $M_s$ and is
logically partitioned into chunks of $\Delta$ tokens (the last chunk
may be shorter):
\[
  C_k = \bigl(y_{(k-1)\Delta+1},\dots,y_{\min(k\Delta,T)}\bigr),
  \qquad k=1,\dots,K,\ K=\lceil T/\Delta\rceil.
\]
For any judge query, the verifier emits a single token from
$\{\textsc{Correct},\textsc{Incorrect}\}$. We never sample: instead we
read the verifier logits $\ell_C,\ell_I$ at the judgment position and
form a calibrated \emph{two-way confidence}
\begin{equation}
  p_{\mathrm{2w}}^{\textsc{Correct}}
  \;=\;\frac{\exp(\ell_C)}{\exp(\ell_C)+\exp(\ell_I)},
  \qquad
  p_{\mathrm{2w}}^{\textsc{Incorrect}} = 1 - p_{\mathrm{2w}}^{\textsc{Correct}}.
  \label{eq:p2way}
\end{equation}
The raw verdict is $\arg\max(\ell_C,\ell_I)$; if a confidence threshold
$\tau$ is supplied, a raw \textsc{Correct} is downgraded to
\textsc{Incorrect} whenever $p_{\mathrm{2w}}^{\textsc{Correct}}<\tau$.

\subsection{Hyperparameters and the configurations used in our experiments}
\label{app:algo-hparams}

Table~\ref{tab:hover-hparams} lists every \sys{} hyperparameter together
with the exact values used by the two model pairs in the main paper:
the \emph{Qwen-only} pair (\texttt{Qwen3-235B-A22B-FP8} target,
\texttt{Qwen3-8B} draft) and the \emph{cross-family} pair
(\texttt{GLM-4.7-FP8} target, \texttt{Qwen3.5-9B} draft). The two
configurations differ deliberately: the calibration of the judge logits
depends on the model family. The Qwen judge produces sharply peaked
\textsc{Correct}/\textsc{Incorrect} logits, so we can run with strict
thresholds ($\tau_F=0.998$, $\tau_P=0.985$); the GLM judge is
consistently less confident on the same questions, so we lower every
threshold by roughly $0.05$--$0.1$. Both pairs use the same chunk size
$\Delta=40$ tokens, which we found gives a good balance between
locating the first error finely and amortising the partial-verify
forward.

\renewcommand{\arraystretch}{1.15}
\setlength{\tabcolsep}{3pt}
\begin{table}[h]
\centering
\caption{All \sys{} runtime hyperparameters with the values used in our two
main experiments. ``Sym.'' is the symbol used in
Algorithm~\ref{alg:hover-detailed}; ``Flag'' is the corresponding
command-line argument exposed by \texttt{sglang.launch\_server}.}
\label{tab:hover-hparams}
\scriptsize
\begin{tabular}{@{}llrrl@{}}
\toprule
Sym. & Flag & \makecell{Qwen3-235B\\+ Qwen3-8B} & \makecell{GLM-4.7\\+ Qwen3.5-9B} & Role \\
\midrule
$\Delta$ & \texttt{--hover-chunk-step-tokens} & $40$ & $40$ & chunk size for partial verify \\
$\tau_F$ & \texttt{--hover-full-confidence-threshold} & $0.998$ & $0.95$ & strict full-verify threshold \\
$\tau_F^{\mathrm{rx}}$ & \texttt{--hover-relaxed-full-threshold} & $0.95$ & $0.90$ & relaxed full-verify threshold \\
$\tau_C^{\mathrm{rx}}$ & \texttt{--hover-relaxed-chunk-threshold} & $0.90$ & $0.83$ & relaxed per-chunk threshold \\
$\tau_P$ & \texttt{--hover-partial-confidence-threshold} & $0.985$ & $0.88$ & strict per-chunk threshold \\
$\rho$ & \texttt{--hover-partial-threshold-proportion} & $0.30$ & $0.20$ & frac.\ of \textsc{Incorrect} chunks $\Rightarrow$ reject \\
$\kappa$ & \texttt{--hover-partial-consecutive-threshold} & $2$ & $2$ & trailing \textsc{Incorrect} run $\Rightarrow$ reject \\
$\eta$ & \texttt{--hover-keep-n-correct-chunks} & $0.0$ & $0.0$ & rollback margin (in chunks) \\
$K^{\mathrm{sd}}$ & \texttt{--hover-short-draft-max-chunks} & $2$ & $2$ & short-draft fast-accept ceiling \\
$\tau_{\mathrm{sd}}$ & \texttt{--hover-short-draft-confidence} & $\tau_F^{\mathrm{rx}}$ & $\tau_F^{\mathrm{rx}}$ & short-draft base threshold \\
$T_p$ & \texttt{--hover-premature-partial-verify} & $300$ & $300$ & premature PV trigger (in tokens) \\
$\rho_p$ & \texttt{--hover-premature-threshold-proportion} & $0.20$ & $0.30$ & premature PV \textsc{Incorrect} fraction \\
$B_b,B_s$ & \texttt{--hover-\{target,draft\}-max-total-tokens} & \multicolumn{2}{c}{$5000$ each} & per-rank KV budgets for $M_b,M_s$ \\
\midrule
\multicolumn{5}{l}{\emph{Optional EAGLE3 acceleration (Qwen-only run; both heads use $3$ steps, top-$5$, $32$ draft tokens):}} \\
\multicolumn{2}{l}{\scriptsize\texttt{--hover-draft-eagle-model-path}} & \multicolumn{3}{l}{\scriptsize\texttt{Tengyunw/qwen3\_8b\_eagle3}} \\
\multicolumn{2}{l}{\scriptsize\texttt{--hover-target-eagle-model-path}} & \multicolumn{3}{l}{\makecell[l]{\scriptsize\texttt{lmsys/SGLang-EAGLE3-Qwen3-235B-}\\\scriptsize\texttt{A22B-Instruct-2507-SpecForge-Meituan}}} \\
\bottomrule
\end{tabular}
\end{table}

\noindent
The shifts from the Qwen column to the GLM column have a single source:
GLM's verifier head is less polarised than Qwen's, so the same
factually correct draft yields a lower
$p_{\mathrm{2w}}^{\textsc{Correct}}$. Keeping the Qwen thresholds on
GLM would force essentially every draft into a target restart and
erase the speedup. We therefore re-fit each threshold by examining the
distribution of $p_{\mathrm{2w}}^{F}$ on a small calibration set,
preserving the same accept/reject ratio across the two model pairs.
All other knobs ($\Delta=40$, $\eta=0$, $K^{\mathrm{sd}}=2$, $\kappa=2$,
$T_p=300$) are kept at the same values across both runs.

\subsection{Sub-routines}
\label{app:algo-subroutines}

\paragraph{Holistic (full) verify.}
Build a single judge prompt
$\pi_F(q,y_s)=\textsc{SysFull}\,\Vert\,\textsc{Rubric}\,\Vert\,
\textsc{User}(q,y_s)$ using the \emph{full-verify} prompt template
matched to the task category (math, science multiple-choice, coding,
open-ended, extraction, reasoning). Run one forward pass of $J$ on
$\pi_F$, take the logits $\ell_C,\ell_I$ at the assistant-turn slot,
and return the calibrated verdict from~\eqref{eq:p2way} subject to the
threshold $\tau_F$.

\paragraph{Parallel partial verify (single forward pass).}
Pack $K$ judge queries into one packed prompt sharing a common
prefix, where the $k$-th query asks the judge to evaluate the prefix
$y_{1:k\Delta}$ against the same rubric. Concretely, we concatenate
\[
  \pi_P \;=\; \textsc{SysPartial}\,\Vert\,\textsc{Rubric}\,\Vert\,
            \textsc{Question}(q)\,\Vert\,
            \bigl(C_k \,\Vert\, \textsc{End}\bigr)_{k=1}^{K},
\]
where $\textsc{End}$ is the ChatML close-and-reopen marker
\verb|<|\verb|im_end|\verb|>|\verb|<|\verb|im_start|\verb|>assistant\n|.
A custom Triton attention mask forces each $\textsc{End}$ slot to
attend only to the question prefix plus its own prefix $C_{1:k}$, and
never to later chunks. One forward pass of $J$ therefore yields $K$
independent $\bigl(\ell_C^{(k)},\ell_I^{(k)}\bigr)$ pairs at the cost
of a single prefix-shared decode. From these we obtain per-chunk
verdicts $v_k$ under threshold $\tau_P$, and a last-correct chunk
index $k^\star$ defined as the largest $k$ such that $v_k$ is
\textsc{Correct} and no \textsc{Incorrect} appears earlier; the chunk
run as a whole rejects the draft when either the fraction of
\textsc{Incorrect} chunks exceeds $\rho$ or the trailing run of
\textsc{Incorrect} chunks reaches $\kappa$.

\paragraph{Adopted-prefix length.}
Given $k^\star$, the number of draft tokens we keep before handing
control to $M_b$ is
\begin{equation}
  L^\star \;=\; \min\!\Bigl(T,\ \Delta\cdot\bigl\lfloor
                 \max\!\bigl(0,\,k^\star + 1 - \eta\bigr)\bigr\rfloor\Bigr),
  \label{eq:adopted}
\end{equation}
i.e.\ we roll back $\eta$ chunks from the last correct boundary
(both configurations use $\eta=0$, so the adopted prefix ends exactly
at the last \textsc{Correct} chunk boundary). $L^\star=0$ collapses to
a full restart from $q$.

\subsection{Full pipeline}
\label{app:algo-pipeline}

Algorithm~\ref{alg:hover-detailed} shows the full procedure. After the
optional premature-PV early-exit, every request flows through three
gates: a strict full-verify accept at $\tau_F$, a short-draft
fast-accept that skips partial verify when its overhead would
dominate, and the parallel partial-verify pass — combined with the
relaxed thresholds $(\tau_F^{\mathrm{rx}},\tau_C^{\mathrm{rx}})$ — that
labels each request \textsc{Sm} (ship the draft), \textsc{Sm+Lg}
(\emph{continue} from the longest accepted prefix), or \textsc{Lg}
(\emph{restart} from $q$).

\begin{algorithm}[H]
\small
\DontPrintSemicolon
\caption{\sys{} in detail: draft $\to$ premature PV $\to$ full verify $\to$ short-draft / partial verify / restart}
\label{alg:hover-detailed}
\KwIn{Prompt $q$; small model $M_s$; large model $M_b$ (also used as judge $J$); hyperparameters from Table~\ref{tab:hover-hparams}; max output length $T_{\max}$.}
\KwOut{Final answer $y^\star$ and a fulfilment label $f\in\{\textsc{Sm},\textsc{Sm+Lg},\textsc{Lg}\}$.}
\gcomment{Stage 1: draft $y_s$ with $M_s$, optionally short-circuiting at $T_p$ tokens if a quick PV already shows divergence}
\eIf{$T_p>0$ \textbf{and} $T_p<T_{\max}$ \tcp*[r]{premature-PV early-exit enabled}}{
   $y^{(1)}_s \gets M_s\bigl(q;\ \mathrm{max\_new}=T_p\bigr)$\;
   \If{$M_s$ stopped early (EOS or stop string)}{
       $y_s \gets y^{(1)}_s$;\ \textbf{goto} Stage~2 \ginline{full draft already produced — nothing to early-exit on}
   }
   $K' \gets \lceil |y^{(1)}_s|/\Delta\rceil$\;
   $(v_k)_{k=1}^{K'} \gets \textsc{PartialVerify}(q,y^{(1)}_s;\Delta,\tau_P)$ \ginline{one packed forward judges every prefix at once}
   evaluate the $(\rho_p,\kappa)$ rules on $(v_k)_{k=1}^{K'}$ to obtain $(V_p,k^\star_p)$\;
   \eIf{$V_p=\textsc{Incorrect}$ \textbf{or} $\#\{k:p_{\mathrm{2w}}^{\textsc{Correct},(k)}<\tau_C^{\mathrm{rx}}\}\ge 3$}{
       $L^\star\gets$ Eq.~\eqref{eq:adopted} on $(v_k)$;\ \textbf{goto} Stage~6 \ginline{abort drafting: $M_b$ takes over from $L^\star$}
   }{
       $y_s \gets$ extend $y^{(1)}_s$ with $M_s\bigl(q\,\Vert\,y^{(1)}_s;\ \mathrm{max\_new}=T_{\max}-T_p\bigr)$\;
   }
}{
   $y_s \gets M_s(q;\ \mathrm{max\_new}=T_{\max})$\;
}
\gcomment{Stage 2: full-answer judge — one $J$ pass on the whole draft; accept immediately if the verifier is overwhelmingly confident}
$(\ell_C^F,\ell_I^F)\gets J\!\bigl(\pi_F(q,y_s)\bigr)$; \ compute $p_{\mathrm{2w}}^{F}$ via Eq.~\eqref{eq:p2way}\;
$V_F\gets\textsc{Correct}$ \textbf{if} $\ell_C^F\ge\ell_I^F$ \textbf{and} $p_{\mathrm{2w}}^{F}\ge\tau_F$ \textbf{else} \textsc{Incorrect}\;
\If{$V_F=\textsc{Correct}$}{
   \Return $(y_s,\textsc{Sm})$ \ginline{strict accept: ship the draft, no $M_b$ work needed}
}
\gcomment{Stage 3: short-draft fast-accept — for very short drafts, skip partial verify if the full-verify confidence already clears the relaxed bar}
$K\gets\lceil|y_s|/\Delta\rceil$;\ \ $\tau_{\mathrm{sd}}^{(K)}\gets\min\!\bigl(\tau_{\mathrm{sd}}+0.02\,(K-1),\;\tau_F^{\mathrm{rx}}\bigr)$\;
\If{$K\le K^{\mathrm{sd}}$ \textbf{and} $p_{\mathrm{2w}}^{F}\ge\tau_{\mathrm{sd}}^{(K)}$}{
   \Return $(y_s,\textsc{Sm})$ \ginline{ship draft: too short for PV to amortise, judge already says OK}
}
\gcomment{Stage 4: parallel partial verify — one packed $J$ pass scores every prefix $y_{1:k\Delta}$ and identifies the last \textsc{Correct} chunk $k^\star$}
$(v_k)_{k=1}^{K}\gets\textsc{PartialVerify}(q,y_s;\Delta,\tau_P)$ \ginline{$K$ prefix verdicts at the cost of one decode}
evaluate the $(\rho,\kappa)$ rules on $(v_k)_{k=1}^{K}$ to obtain $(V_P,k^\star)$\;
$p^{\min}_{\mathrm{2w}}\gets\min_{k} p_{\mathrm{2w}}^{\textsc{Correct},(k)}$\;
\gcomment{Stage 5: relaxed accept — full verdict missed strict bar, but if $J$ is moderately confident overall AND every chunk passed locally, ship it}
\If{$p_{\mathrm{2w}}^{F}\ge\tau_F^{\mathrm{rx}}$ \textbf{and} $p^{\min}_{\mathrm{2w}}\ge\tau_C^{\mathrm{rx}}$}{
   \Return $(y_s,\textsc{Sm})$ \ginline{ship draft: globally OK + locally OK on every chunk}
}
\gcomment{Stage 6: hand off to $M_b$ — \textbf{continue} from the longest accepted prefix $y_{1:L^\star}$ if any, otherwise \textbf{restart} from the prompt $q$}
$L^\star\gets$ Eq.~\eqref{eq:adopted} with $k^\star$ and $\eta$\;
\eIf{$L^\star>0$}{
   $y^\star\gets y_{1:L^\star}\,\Vert\,M_b\bigl(q\,\Vert\,y_{1:L^\star}\bigr)$ \ginline{\textbf{continue}: reuse $L^\star$ correct draft tokens, $M_b$ writes only the suffix}
   \Return $(y^\star,\textsc{Sm+Lg})$\;
}{
   $y^\star\gets M_b(q)$ \ginline{\textbf{restart}: no usable prefix, $M_b$ writes the full answer}
   \Return $(y^\star,\textsc{Lg})$\;
}
\end{algorithm}


\section{SpecReason baseline}
\label{appendix:specreason-config}

The SpecReason baseline uses the same Qwen3-235B / Qwen3-8B target /
drafter pair as our \sys{} runs, with a fixed chunk size of
$\Delta_{\mathrm{SR}} = 40$ tokens. Algorithm~\ref{alg:specreason-baseline}
gives the per-question control flow.

\begin{algorithm}[H]
\small
\DontPrintSemicolon
\caption{SpecReason baseline (one question)}
\label{alg:specreason-baseline}
\KwIn{Prompt $q$; small server $S_s$; big server $S_b$; score
threshold $\tau{=}7$; chunk size $\Delta_{\mathrm{SR}}{=}40$ tokens;
token budget $B{=}1300$; reasoning- or code-scoring prompt $\pi$.}
\KwOut{Final response $y$.}
$y \gets \varepsilon$;\ \ $T \gets 0$\;
\While{$T < B$ \textbf{and} $S_s$ has not emitted EOS}{
   $c_s \gets S_s.\textsc{Generate}(q,\,y;\,\Delta_{\mathrm{SR}})$ \ginline{small drafter proposes one chunk}
   $s \gets S_b.\textsc{ScoreFirstDigit}\!\bigl(q,\,y,\,c_s,\,\pi\bigr)$ \ginline{single-token logprob query on $S_b$}
   \eIf{$s \ge \tau$}{
      $y \gets y \,\Vert\, c_s$ \ginline{accept: cheap chunk}
   }{
      $c_b \gets S_b.\textsc{Generate}(q,\,y;\,\Delta_{\mathrm{SR}})$ \ginline{regenerate the rejected chunk on $S_b$}
      $y \gets y \,\Vert\, c_b$\;
   }
   $T \gets$ $|y|$ in tokens\;
}
\Return $y$\;
\end{algorithm}

\section{Prompt for Stage 2 of Algorithm~\ref{alg:draft-rescue}}

In Stage~2(i) (\emph{Full-answer judgement}), the target model holistically evaluates whether the draft answer is correct (see Algorithm~\ref{alg:draft-rescue}, line~4).
This verification is conducted in a \emph{prefill-only} manner: the entire draft is passed once through the model, which then emits a classification token (\texttt{Correct} or \texttt{Incorrect}).
The following prompt specifies the verification instruction used by the model in this stage.

\begin{tcolorbox}[breakable, colback=white,colframe=black!50,title=Stage 2(i) Verification Prompt,fonttitle=\bfseries]
\textbf{SYSTEM PROMPT} \\
You are a cold, conservative, strict and rigorous verifier of COMPLETE solutions.
Given a question and a proposed answer, decide if the answer is correct.
There are three modes—pick exactly one:
- Math: math problems/derivations/numeric computations.
- Science-QA: science multiple-choice (A–J) style questions.
- Coding: programming/code-generation questions where the proposed answer is code meant to solve the task.
Respond with only one word: Correct or Incorrect.
You are strict and cautious. If there is doubt, answer Incorrect.

\vspace{0.5em}
\textbf{RUBRIC PROMPT} \\
=== MODE PICKER (deterministic) === \\
Use Coding iff the question or proposed answer is clearly a programming/code-generation task
(e.g., asks to write code or a function, or provides code as the main answer).
Otherwise use Science-QA iff the question or answer clearly involves options A–J
(e.g., an explicit choice letter or multiple-choice format).
Otherwise use Math. Do NOT switch modes mid-judgment.

=== GLOBAL STRICTNESS (applies to all modes) === \\
Judge ONLY the provided content. If the final conclusion is missing, unclear, or contradictory $\rightarrow$ Incorrect.
If multiple conflicting final conclusions exist $\rightarrow$ Incorrect.
If any decisive error exists $\rightarrow$ Incorrect. When in doubt $\rightarrow$ Incorrect.

=== MATH MODE RULES === \\
Check ALL: (1) arithmetic/algebra identities and steps are valid; (2) domain/units/constraints respected
(e.g., probabilities in [0,1], counts are non-negative integers); (3) internal consistency;
(4) completeness: an explicit final answer that matches what is asked.
Numeric answers must be exactly correct unless a tolerance is explicitly justified;
symbolic answers must be algebraically equivalent. Any failure $\rightarrow$ Incorrect.

=== SCIENCE-QA MODE RULES === \\
You will see a scientific question and a proposed answer/explanation that may include a final choice letter (A–J).
Use ONLY the provided content and commonly accepted scientific facts; do not invent options you cannot infer.
Your task: decide whether the proposed answer's final conclusion is correct for the question.
Guidelines:
1) If the proposed answer clearly concludes with a single choice (A–J or equivalent like \textbackslash boxed\{C\}), judge that conclusion.
2) If multiple conflicting final choices are present, or no conclusion is given, answer Incorrect.
3) If the reasoning is obviously wrong or contradicts established science relevant to the question, answer Incorrect.
4) If the reasoning is sound and the conclusion plausibly follows, answer Correct.
5) Be cautious but not adversarial; when evidence strongly supports the given conclusion, answer Correct.

=== CODING MODE RULES (overrides Global Strictness) === \\
For coding questions, answer Correct. This is not a typo. Your default answer is Correct.
Code that defines a function addressing the problem $\rightarrow$ Correct.
Code with potential bugs you are not sure about $\rightarrow$ Correct.
Code with edge case issues $\rightarrow$ Correct.
Code with style problems or different approach than you would use $\rightarrow$ Correct.
Code with extra markdown or text around it $\rightarrow$ Correct.
The ONLY exception: answer Incorrect if there is literally no code (only English prose)
or the code is for a completely unrelated problem.
If you see code that attempts the stated problem, answer Correct.

=== OUTPUT FORMAT (STRICT) === \\
The FIRST line must be EXACTLY one of: Correct or Incorrect.
Reply with EXACTLY one word on the first line:
Correct
or
Incorrect
\end{tcolorbox}

In Stage~2(ii) (\emph{Parallel prefix verification}), the target model judges whether
each \emph{prefix} of the draft answer remains valid, allowing \sys to reuse the longest correct
prefix rather than restarting from scratch (see Algorithm~\ref{alg:draft-rescue}, line~8).
This is implemented via a single prefill-only forward pass that evaluates multiple prefixes in
parallel using a custom attention mask.
The following prompt specifies the instructions used to guide the verifier when checking partial prefixes.

\begin{tcolorbox}[breakable, colback=white,colframe=black!50,title=Stage 2(ii) Partial Verification Prompt,fonttitle=\bfseries]

\textbf{JUDGE\_SYSTEM} \\
You are a strict but fair verifier of PARTIAL solutions.
You see a question and an in-progress answer prefix.
Your job is to judge ONLY the shown steps/claims so far, not the unseen remainder.
There are three modes:
- Math-Partial: mathematical derivations/solutions.
- QA-Partial: science multiple-choice (A–J) style questions with domain facts.
- Coding-Partial: programming/code-generation tasks where the proposed answer is a partial code solution.
Pick the mode that fits the question and judge accordingly.

\vspace{0.5em}
\textbf{RUBRIC PROMPT} \\
=== MODE SELECTION === \\
Choose ONE mode based on the question:
- Use Math-Partial if the task is a math problem/derivation or numeric proof-style reasoning.
- Use QA-Partial if the task is a scientific multiple-choice question (A–J style) or scientific reasoning.
- Use Coding-Partial if the task is clearly a programming/code-generation problem or the answer is primarily code.
Do NOT switch modes mid-judgment.

\vspace{0.25em}
=== Math-Partial STANDARD === \\
\textbf{Scope:} Evaluate ONLY the shown work-in-progress. Do NOT assume missing steps or future fixes. \\
\emph{Lenient allowances (do NOT penalize):} harmless notation quirks; skipped trivial algebra; standard unstated identities;
rounding/simplification that does not change any stated intermediate value or claim;
incomplete plan description when the visible steps remain consistent and legal. \\
\emph{Strict errors (mark Incorrect):} arithmetic/algebra mistakes that change a stated value; invalid identities;
illegal operations (e.g., divide by 0); domain or constraint violations
(e.g., probabilities outside [0,1], negative lengths declared positive);
logical contradictions within the shown steps; misuse of definitions or theorems yielding a wrong derived claim. \\
\emph{Tie-breaker:} If you cannot point to a concrete mathematical error in the SHOWN steps, output Correct.

\vspace{0.25em}
=== QA-Partial STANDARD === \\
\textbf{Scope:} Evaluate ONLY the shown scientific statements/inferences relative to the question stem (A–J style context).
You are checking whether the partial reasoning already contains a decisive scientific or quantitative mistake. \\
\emph{Lenient allowances (do NOT penalize):} lack of a final option letter; exploratory or incomplete reasoning that
remains consistent with the stem; benign phrasing or notation differences; standard high-level summaries of well-known facts. \\
\emph{Strict errors (mark Incorrect):} factual falsehoods about established science relevant to the question
(e.g., mis-stated definitions/laws, wrong matrix or operator forms, impossible units/dimensions);
quantitative or linear-algebra mistakes that change a stated intermediate claim
(e.g., wrong eigenvalue/eigenvector relation, invalid normalization);
inferences that contradict given conditions or data in the stem, or internal contradictions within the shown reasoning;
explicit commitment to an option that is logically ruled out by the stem or by the shown calculations. \\
\emph{Neutral/incomplete:} If the partial text has not yet made a falsifiable claim and contains no concrete error, output Correct. \\
\emph{Tie-breaker:} Absent a definite scientific or quantitative error in the SHOWN text, output Correct.

\vspace{0.25em}
=== Coding-Partial STANDARD === \\
\textbf{Scope:} Evaluate ONLY the visible portion of the code/solution, not any imagined future edits. \\
\textbf{Default:} For code that attempts the stated problem, your default answer is Correct. \\
\emph{Lenient allowances (do NOT penalize — answer Correct):} missing trailing code or obvious TODOs when the existing
code remains logically consistent; benign style issues, minor inefficiencies, or different variable/parameter names;
different algorithmic approach than you would use, as long as the visible logic is consistent;
code that uses different built-in functions or standard library calls than you would choose;
partial or incomplete code that could still be extended to a correct solution. \\
\emph{Strict errors (mark Incorrect ONLY for these):} clear logic bugs that would definitely produce wrong results on any input;
code that is for a completely unrelated problem or contains no code at all. \\
\emph{Tie-breaker:} When in doubt about code correctness, output Correct.

\vspace{0.25em}
=== OUTPUT FORMAT (STRICT) === \\
Reply with EXACTLY one word on the first line:
Correct
or
Incorrect
No other text is allowed.

\end{tcolorbox}

\paragraph{Limitation and scope of the prompts.}
The two judge prompts above are deliberately domain-aware. We acknowledge that this domain-specific
prompt engineering is a limitation --- a different deployment may
need a fresh rubric for its own task mix, and the reported numbers
should be read as evidence that \emph{some} good prompt exists,
rather than as a one-size-fits-all artifact. The contribution of
this paper, however, is not the particular wording of the rubric.
It is the higher-level claim that an LLM target model already
\emph{knows when its own draft has gone wrong}, and that this
knowledge can be extracted in parallel across every prefix of the
draft in a single forward pass --- enabling cheap, fine-grained
error localisation rather than full regeneration. We welcome and
encourage further prompt-engineering work that strengthens the
verifier on top of this primitive.

\section{Examples of Continual Generation from the last correct prefix}
In this section, we provide illustrative case studies showing how \sys leverages prefix verification to recover from draft model errors. Specifically, when the draft model deviates from the correct solution, the target model is able to identify the earliest point of error and resume generation from the last valid prefix, rather than restarting from scratch. These examples highlight the effectiveness of continual generation: by reusing correct portions of the draft, \sys minimizes redundant computation while still ensuring accurate final answers. We follow the same convention as in section~\ref{case_study}, where black text is the reused prefix, red text is the rejected portion of the draft, and the green text is the new suffix regenerated by the target model.

\HRule

\noindent\textbf{Question:} The largest exporter of agricultural goods in the world is :A. the United States.B. Mexico. C. Chile. D. China?

\HRule

To determine the largest exporter of agricultural goods in the world, we need to consider the global agricultural trade landscape. 1. United States (A): The U.S. is a major agricultural producer and exporter, particularly in crops like corn, soybeans, and wheat. However, it is not the largest exporter globally in terms of total value. 2. Mexico (B): Mexico is a major exporter of agricultural goods, especially in fruits, vegetables, and dairy products. It is one of the largest agricultural exporters in the world, but it is not the largest. 3. Chile (C): Chile is \textcolor{red}{a major exporter of agricultural goods, particularly in wine, beef, and other high-value products. It is one of the top agricultural exporters in the world. 4. China (D): China is the largest exporter of agricultural goods in the world, with significant exports of rice, wheat, and other staple crops. It is the largest in terms of total value and volume. Final Answer: D} \textcolor{ForestGreen}{a significant exporter of fruits, wine, and seafood, particularly to North America and Asia. 4. China (D): China is the world’s largest agricultural producer, but it is not a top net exporter. In fact, the United States consistently ranks as the largest exporter of agricultural goods by total value, according to data from the USDA and the World Trade Organization. While other countries may lead in specific commodities, the U.S. leads in overall agricultural export value. Final Answer: A
}

\HRule

\noindent\textbf{Question:} A custodian has to clean a school with 80 classrooms. They have 5 days to get it done. It takes them 15 minutes per classroom. If they work an 8 hour day, what percentage of their day, on average, is spent cleaning classrooms?

\HRule

To find the percentage of the custodian's day spent cleaning classrooms, we need to follow these steps: Step 1: Total time spent cleaning classrooms - There are 80 classrooms. - It takes 15 minutes per classroom. - Total time spent cleaning = $80 \times 15 = 1200 \text{ minutes}$. Step 2: Convert total time to hours - $1200 \text{ minutes} = \tfrac{1200}{60} = 20 \text{ hours}$. \textcolor{red}{Step 3: Total time available in a day - The custodian works $8$ hours per day. Step 4: Calculate the percentage of the day spent cleaning - $\text{Percentage} = \left(\tfrac{20}{8}\right) \times 100 = 250\%$. Final Answer: $250\%$} \textcolor{ForestGreen}{Step 3: Total available work hours - The custodian works 8 hours per day for 5 days. - Total available work time = $8 \times 5 = 40 \text{ hours}$. Step 4: Calculate percentage of time spent cleaning - Time spent cleaning = $20$ hours. - Total work time = $40$ hours. - Percentage of time spent cleaning = $\left(\tfrac{20}{40}\right) \times 100 = 50\%$. The custodian spends 50\% of their time, on average, cleaning classrooms. Final Answer: $50\%$
}

\HRule

\noindent\textbf{Question:} There are 4 snails in one aquarium and 32 snails in another aquarium. The difference between the number of snails in the two aquariums is twice the amount of fish in both aquariums. If both aquariums have the same number of fish in them, how many fish are there in each aquarium?

\HRule

We are given the following information: - There are 4 snails in one aquarium. - There are 32 snails in another aquarium. - The difference between the number of snails in the two aquariums is twice the amount of fish in both aquariums. - Both aquariums have the same number of fish. Step 1: Find the difference in the number of snails $32 - 4 = 28$. So, the difference in the number of snails is 28. Step 2: Use the given relationship We are told that this difference (28) is twice the amount of fish in both aquariums. Let the number of fish in \textcolor{red}{each aquarium be $x$. Then, the total number of fish in both aquariums is: $x + x = 2x$. We are told: $2x = 28$. Step 3: Solve for $x$ $x = \tfrac{28}{2} = 14$. Final Answer: $14$} \textcolor{ForestGreen}{each aquarium be $f$. Since both aquariums have the same number of fish, the total number of fish in both aquariums is: $f + f = 2f$. According to the problem, the difference in snails (28) is twice the total number of fish in both aquariums: $28 = 2 \times (2f)$. Simplify the right side: $28 = 4f$. Now solve for $f$: $f = \frac{28}{4} = 7$. Final Answer: $7$. There are 7 fish in each aquarium.
}

\HRule

\noindent\textbf{Question:} What is the traditional 20th wedding anniversary gift? Options: A. silver B. emerald C. pearl D. diamond E. leather F. sapphire G. ruby H. paper I. china J. gold

\HRule

To determine the traditional 20th wedding anniversary gift, we need to refer to the traditional gift associations for wedding anniversaries. Here is the traditional gift list for wedding anniversaries: - 1st: Paper - 2nd: Cotton - 3rd: Leather - 4th: Fruit - 5th: Wood - 6th: Iron - 7th: Wool - 8th: Cotton - 9th: Cotton - 10th: Tin - 11 \textcolor{red}{th: Copper - 12th: Fabric - 13th: Fabric - 14th: Crystal - 15th: Crystal - 16th: Lapis Lazuli - 17th: Lapis Lazuli - 18th: China - 19th: China - 20th: Silver. So, the traditional 20th wedding anniversary gift is silver. Final Answer: $\boxed{A}$}  \textcolor{ForestGreen}{11th: Steel - 12th: Silk - 13th: Lace - 14th: Ivory - 15th: Crystal - 20th: China. The traditional 20th wedding anniversary gift is china. $\boxed{I}$
}

\HRule

\noindent\textbf{Question:} What is the largest order of an element in the group of permutations of 5 objects? Options: A. 12 B. 6 C. 5 D. 15

\HRule

To determine the largest order of an element in the group of permutations of 5 objects, we need to understand the structure of the symmetric group $S_5$. Step 1: Understand the group The symmetric group $S_n$ consists of all permutations of $n$ objects. The order of a permutation is the least common multiple (LCM) of the lengths of its disjoint cycles. Step 2: Find the largest possible cycle length In $S_5$, the maximum possible cycle length is 5 (a single 5-cycle). The order of a 5-cycle is 5, since the LCM of 5 is 5. Step 3: Consider other cycle types Other cycle types in $S_5$ include: - 4-cycles (order 4) - 3-cycles (order 3) \textcolor{red}{- 2-cycles (order 2) - 1-cycles (order 1). The largest order among these is 5 (from a 5-cycle). Step 4: Confirm the answer The largest order of an element in $S_5$ is 5. Final Answer: $\boxed{C}$}  \textcolor{ForestGreen}{Products of disjoint cycles, such as a 3-cycle and a 2-cycle (disjoint), which gives a permutation of type (3,2). We must consider all integer partitions of 5 and compute the LCM of the cycle lengths. Partitions of 5 and corresponding LCMs: 1. (5) → LCM(5) = 5 2. (4,1) → LCM(4,1) = 4 3. (3,2) → LCM(3,2) = 6 4. (3,1,1) → LCM(3,1,1) = 3 5. (2,2,1) → LCM(2,2,1) = 2 6. (2,1,1,1) → LCM(2,1,1,1) = 2 7. (1,1,1,1,1) → identity, order 1. The maximum among these is LCM(3,2) = 6. Is there a partition that gives a higher LCM? Try (3,2): 3+2=5 → valid, LCM=6. (5): LCM=5. (4,1): 4. (2,3): same as (3,2). No partition gives LCM higher than 6. Wait: is there a way to get LCM = 12? That would require cycles of length 3 and 4, but 3+4=7 > 5 — impossible. LCM = 15? Needs 3 and 5 → 3+5=8 > 5 — impossible. So maximum is 6. But wait — is there a 5-cycle? Order 5. Or a 3-cycle and a disjoint 2-cycle: yes, in $S_5$, we can have (1 2 3)(4 5). This has order LCM(3,2) = 6. No other combination gives higher. Thus, the largest order is 6. $\boxed{B}$
}

\end{document}